\documentclass[sigconf,nonacm]{acmart}
\usepackage{multirow}
\definecolor{mygreen}{RGB}{0, 153, 0}
\definecolor{myblue}{RGB}{0, 0, 204}
\definecolor{myred}{RGB}{204, 0, 0}

\AtBeginDocument{%
  }

\begin{document}

\title{Beyond Homoscedasticity: Decoupled Uncertainty Optimization for Deep Imbalanced Regression}
\thanks{Accepted at the 34th ACM International Conference on Multimedia
(ACM MM 2026). The Version of Record will be available at
\url{https://doi.org/10.1145/3767308.3835898}.}

\author{Juncheng Zhou}
\authornote{Juncheng Zhou and Jiaxi Lu contributed equally to this work.}
\orcid{0009-0002-2151-9058}
\affiliation{%
  \institution{School of Cyber Science and Engineering, Wuhan University}
  \city{Wuhan}
  \country{China}}
\email{2020302181088@whu.edu.cn}

\author{Jiaxi Lu}
\authornotemark[1]
\orcid{0009-0000-6123-9961}
\affiliation{%
  \institution{School of Cyber Science and Engineering, Wuhan University}
  \city{Wuhan}
  \country{China}}
\email{lulululu@whu.edu.cn}

\author{Weijing Zeng}
\orcid{0009-0009-2463-7675}
\affiliation{%
  \institution{School of Mathematics and Statistics, Wuhan University}
  \city{Wuhan}
  \country{China}}
\email{2021302011066@whu.edu.cn}

\author{Zhong Li}
\orcid{0009-0007-0523-3638}
\affiliation{%
  \institution{School of Synthetic Biology and Biomanufacturing, Tianjin University}
  \city{Tianjin}
  \country{China}}
\email{lz95\_@tju.edu.cn}

\author{Hao Qi}
\orcid{0000-0002-2849-6226}
\affiliation{%
  \institution{School of Synthetic Biology and Biomanufacturing, Tianjin University}
  \city{Tianjin}
  \country{China}}
\email{haoq@tju.edu.cn}

\author{Jingsong Cui}
\correspondingauthor
\orcid{0000-0002-3111-9798}
\affiliation{%
  \institution{School of Cyber Science and Engineering, Wuhan University}
  \city{Wuhan}
  \country{China}}
\email{jscui@whu.edu.cn}

\renewcommand{\shortauthors}{Juncheng Zhou, Jiaxi Lu, Weijing Zeng, Zhong Li, Hao Qi, \& Jingsong Cui}

\begin{abstract}
Deep Imbalanced Regression (DIR) is pervasive in continuous prediction tasks across diverse modalities, such as age estimation, depth prediction, and protein mutation activity prediction, where label-scarce tail samples often carry higher practical value. However, most existing methods still learn deterministic point mappings under mean squared error or its simple variants, implicitly assuming a uniform uncertainty level across all samples and thereby overlooking the instance-wise heteroscedasticity that is widespread in long-tailed data. We further point out that even heteroscedastic negative log-likelihood suffers from a gradient coupling issue, which, under DIR scenarios, weakens the learning signal of hard tail samples and leads to optimization inertia as well as tail underfitting. To address this, we propose DUO, an uncertainty-aware long-tailed regression framework. Specifically, the proposed method models the regression target as a conditional Gaussian distribution to explicitly characterize instance-level predictive uncertainty, and transforms uncertainty into a dynamic enhancement signal for tail samples through decoupled mean-variance optimization. Furthermore, we design a distribution-guided contrastive learning mechanism that adaptively constructs positive and negative pairs based on the overlap between sample distributions, thereby alleviating feature looseness and cross-label semantic entanglement. Across visual and biological DIR benchmarks, DUO achieves the best few-shot bMAE and GM on IMDB-WIKI-DIR, AgeDB-DIR, and AAV2-DIR while remaining competitive on few-shot MAE.
\end{abstract}



\maketitle

\section{Introduction}

\begin{figure}[t]
    \centering
    \includegraphics[width=0.98\columnwidth]{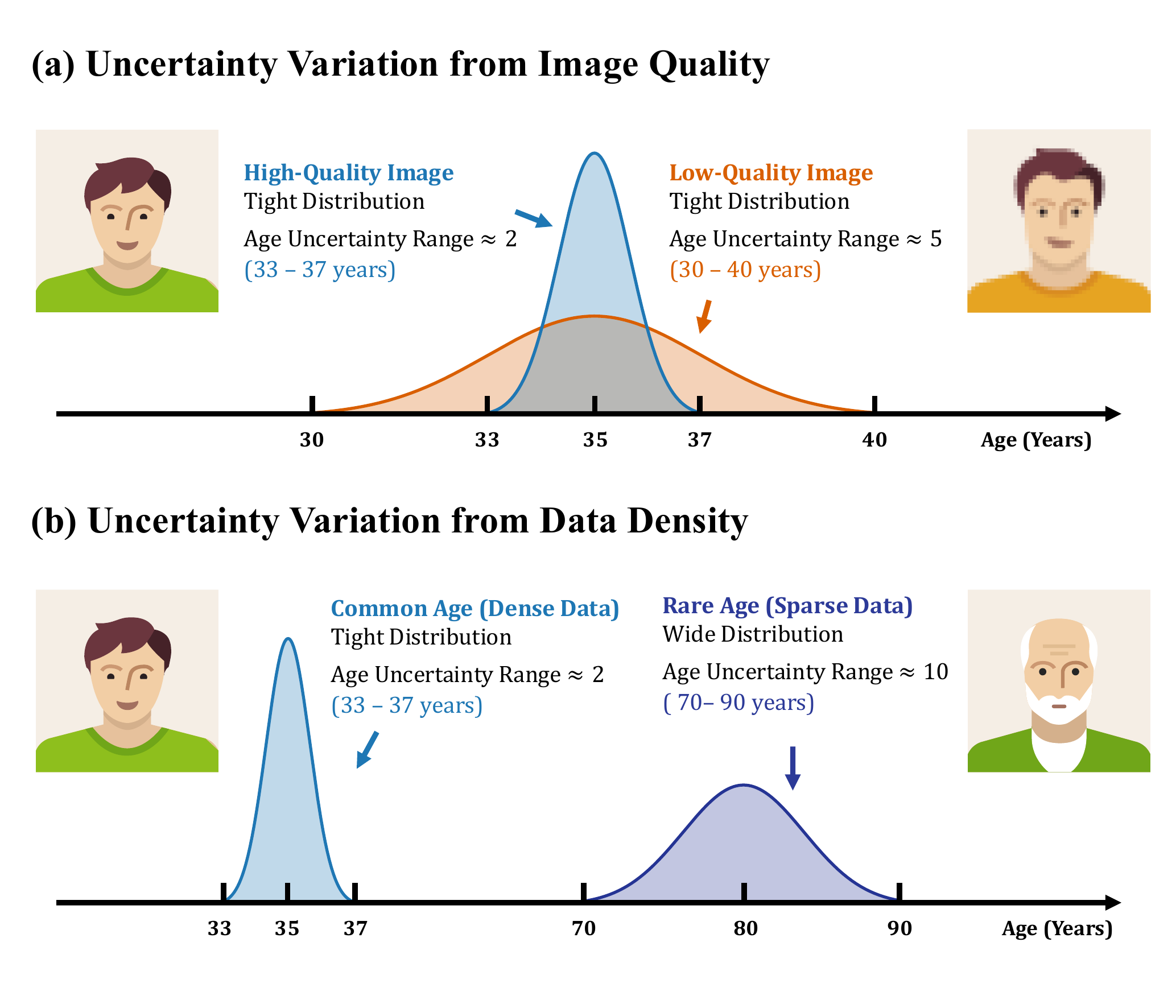}
    \caption{\textbf{Illustration of instance-level heteroscedasticity.} 
(a) For the same age label, predictive uncertainty increases as image quality degrades. 
(b) Predictive uncertainty also increases in sparse label regions, even for high-quality samples.}
    \label{fig:real}
\end{figure}

Deep Imbalanced Regression (DIR) is prevalent across numerous real-world continuous prediction tasks, ranging from facial age and depth estimation in multimedia systems to protein mutation activity prediction in computational biology. Such tasks typically exhibit a prominent long-tailed distribution, wherein a majority of the labels dominate while a minority remain extremely scarce. This data imbalance heavily biases optimization toward the massive, easy-to-fit majority ("Many"). Consequently, models suffer from severe prediction bias against the scarce tail ("Few"), leading to systematically less reliable predictions on underrepresented samples. Tail categories can nevertheless have substantial collective influence\cite{zhang2025systematic}, and individual rare cases may be consequential in application-specific settings.

For instance, Deep Mutational Scanning (DMS)\cite{fowler2014deep,fowler2014measuring} of AAV2 capsid proteins\cite{ogden2019comprehensive} provides large fitness landscapes with pronounced imbalance. Bryant et al.~\cite{bryant2021deep} focus on a 28-residue AAV2 segment (VP1 positions 561--588) that overlaps known antibody-binding sites and generate highly diverse variants that remain viable for packaging. In our processed AAV2-DIR data, the fraction labeled active drops from approximately 10\% to 0.3\% as the mutation count increases ($\geq 6$), approaching zero beyond 21 mutations. This scarcity yields a highly skewed target distribution, making reliable prediction in the sparse region important for identifying unusual viable variants.

To alleviate long-tailed bias, existing research extensively explores data re-weighting, re-sampling, and architecture design\cite{yang2021delving, steininger2021density, torgo2013smote, branco2017smogn, gong2022ranksim}. However, most methods still optimize using Mean Squared Error (MSE) or its simple variants as the regression loss, which limits their ability to reflect sample-wise uncertainty. From a statistical modeling perspective, MSE entails an implicit assumption of global homoscedasticity, which presumes that all samples share an identical level of confidence (or uncertainty) during prediction.

Real-world data often exhibit instance-level heteroscedasticity, where predictive uncertainty varies with both image quality and label density, which is particularly relevant to responsible multimedia systems that require reliable predictions under heterogeneous conditions. As shown in Figure~\ref{fig:real}(a), even at the same age label (e.g., 35 years old), a clear image yields a sharp distribution with low uncertainty (33-37 years, $\sigma \approx 2$), whereas a blurry image produces a much wider distribution (30-40 years, $\sigma \approx 5$). Figure~\ref{fig:real}(b) further shows that uncertainty also depends on label density: a common-age sample can remain highly confident (33-37 years, $\sigma \approx 2$), while a high-quality sample from a sparse tail region (e.g., age 80) may still exhibit much higher uncertainty (70-90 years, $\sigma \approx 10$). These observations reveal the limitation of MSE, which implicitly assumes uniform uncertainty across samples and may obscure reliability disparities across underrepresented cases.

While mainstream DIR methods have made significant progress, their reliance on homoscedastic assumptions often introduces two distinct challenges. At the optimization level, highly uncertain tail samples can disrupt gradient updates, potentially leading to underfitting in scarce regions and reduced reliability on rare samples. At the representation level, treating heterogeneous samples uniformly tends to entangle cross-label semantics and degrade intra-label feature compactness. Consequently, shifting from deterministic point mapping to modeling samples as conditional probability distributions has emerged as a promising paradigm to capture data heterogeneity and provide more transparent uncertainty estimates.

Following this paradigm, prior works have explored heteroscedastic modeling using the standard Negative Log-Likelihood (NLL) loss. However, jointly optimizing the mean and variance in NLL can compromise mean fitting (Stirn et al. \cite{stirn2023faithful}). Under long-tailed distributions, this issue is drastically exacerbated, inducing "optimization inertia." Recent uncertainty-aware DIR approaches instead adopt probabilistic smoothing or multi-expert aggregation\cite{wang2023variational,jiang2024uncertainty}, but these mechanisms add modeling or inference complexity.

To address these challenges, we propose DUO, an uncertainty-guided long-tailed regression framework. By modeling the prediction target as a conditional Gaussian distribution, DUO explicitly captures instance-level uncertainty through two core designs, thereby better characterizing sample-wise predictive reliability. First, a mean-variance decoupled optimization mechanism utilizes gradient detachment to adaptively weight gradients, mitigating the gradient degradation issue in traditional likelihood losses. Second, a distribution-guided contrastive learning module employs the Bhattacharyya coefficient to quantify distribution overlap, thereby adaptively constructing contrastive pairs to alleviate the feature looseness and semantic entanglement of tail samples.

In summary, our main contributions are threefold: (1) we formulate long-tailed regression as a conditional Gaussian problem to capture instance-level uncertainty, and analyze how mean--variance gradient coupling in NLL impairs optimization in DIR; (2) we propose the DUO loss function, which incorporates gradient detachment for computationally decoupled optimization and distribution-guided contrastive learning; and (3) we validate DUO across multimedia and biological DIR benchmarks, demonstrating consistent gains on imbalance-aware few-shot metrics.

\section{Related Work}
\subsection{Imbalanced Classification}
Existing imbalanced classification methods are broadly divided into two mainstream paradigms: reweighting/resampling-based approaches (RW/RS) and decision boundary and prior calibration-based approaches (DBC/PC). RW/RS methods\cite{chawla2002smote,han2005borderline,jiang2021improving} mitigate long-tailed optimization bias by adjusting the gradient contributions of different classes/samples during training, with representative works including cost-sensitive learning based on inverse class frequency, Class-Balanced Loss that avoids over-amplification of tail classes via effective sample number (Cui et al.\cite{cui2019class}), and Focal Loss that alleviates gradient domination by focusing on hard samples (Lin et al.\cite{lin2017focal}). DBC/PC methods\cite{zhong2021improving,yin2019feature} tackle imbalance from the perspective of classification boundary geometry or label prior mismatch, such as LDAM with class-dependent margin optimization (Cao et al.\cite{cao2019learning}), the decoupled training paradigm that separates representation learning and classifier rebalancing (Kang et al.\cite{kang2019decoupling}), and Logit Adjustment(Menon et al.\cite{menon2020long}) and Balanced Softmax(Ren et al.\cite{ren2020balanced}) that correct prior mismatch at the logit level.

However, all above methods are built on the structural assumptions of discrete classes, class priors, or classification boundaries. This inherent discrete label dependency makes them difficult to directly transfer to imbalanced regression tasks with continuous label spaces, where the coupling between sample scarcity, predictive uncertainty and feature geometry remains to be systematically modeled.

\subsection{Imbalanced Regression}
Compared with imbalanced classification, imbalanced regression faces greater challenges due to the continuous nature of label space and the absence of explicit class boundaries. Sample scarcity in IR no longer manifests as class frequency imbalance, but as highly skewed label distributions with severe sparsity and noise in tail regions (Yang et al.\cite{yang2021delving}), making most classification methods inapplicable. Existing IR approaches typically intervene at the input, feature, or output level, but most operate on only a single level and fail to model the coupling between sample scarcity and predictive uncertainty.

\textbf{Input-level methods} mitigate label imbalance by adjusting training data distribution or sample weights directly\cite{chawla2002smote,torgo2013smote,branco2017smogn,branco2018rebagg}. Early works like SMOTER\cite{torgo2013smote} and SMOGN\cite{branco2017smogn} define rare regions in continuous label space, and perform oversampling for tail samples while undersampling head ones. However, these methods only adjust the data distribution at the input stage, without addressing the core issues of feature collapse and prediction bias in tail regions during model optimization.

\textbf{Feature-level methods} preserve continuous label relationships in representation space via structural constraints on the feature space. The seminal DIR work proposes Label Distribution Smoothing (LDS) and Feature Distribution Smoothing (FDS), which alleviate tail feature collapse by sharing statistical information within local label neighborhoods (Yang et al.\cite{yang2021delving}). Building on this, RankSim\cite{gong2022ranksim} introduces global geometric constraints by aligning the relative ordering between label and feature spaces, and ConR\cite{keramati2023conr} further leverages contrastive learning to penalize label-feature inconsistency and prevent minority sample collapse. While these methods effectively model feature geometry for continuous labels, they do not incorporate predictive uncertainty into the feature learning process, leaving the coupling between sample scarcity and uncertainty under-explored.

\textbf{Output-level methods} mainly mitigate tail prediction bias by modifying loss functions or predictive distributions. DenseWeight\cite{steininger2021density} estimates target rarity via kernel density estimation, while LDS-based reweighting smooths empirical label densities before assigning sample weights\cite{yang2021delving}. Balanced MSE (BMSE)\cite{ren2022balanced} corrects the discrepancy between training and target priors, and Dist Loss\cite{nie2024dist} jointly penalizes sample-wise errors and prediction--label distribution distance. These methods correct prediction bias but do not use uncertainty to guide upstream representation learning.

Recent works have introduced uncertainty into imbalanced regression: VIR~\cite{wang2023variational} uses probabilistic neighborhood smoothing for long-tailed uncertainty quantification, while UVOTE~\cite{jiang2024uncertainty} uses predictive uncertainty to select among ensemble experts. These methods model uncertainty primarily at the output or predictive-distribution level rather than using it to construct representation-space relations, leaving uncertainty-guided feature learning underexplored.

In summary, existing IR methods generally alleviate label sparsity-induced bias from a single perspective, but fail to explicitly model the coupling among sample scarcity, predictive uncertainty, and feature geometry in continuous label spaces. Motivated by this gap, we propose \textbf{DUO}, which explicitly models sample-level uncertainty and uses it as a unified guidance signal for both optimization and representation learning, thereby improving tail sample modeling in long-tailed regression.

\section{Method}

\begin{figure*}[t]
    \centering
    \includegraphics[width=1\textwidth]{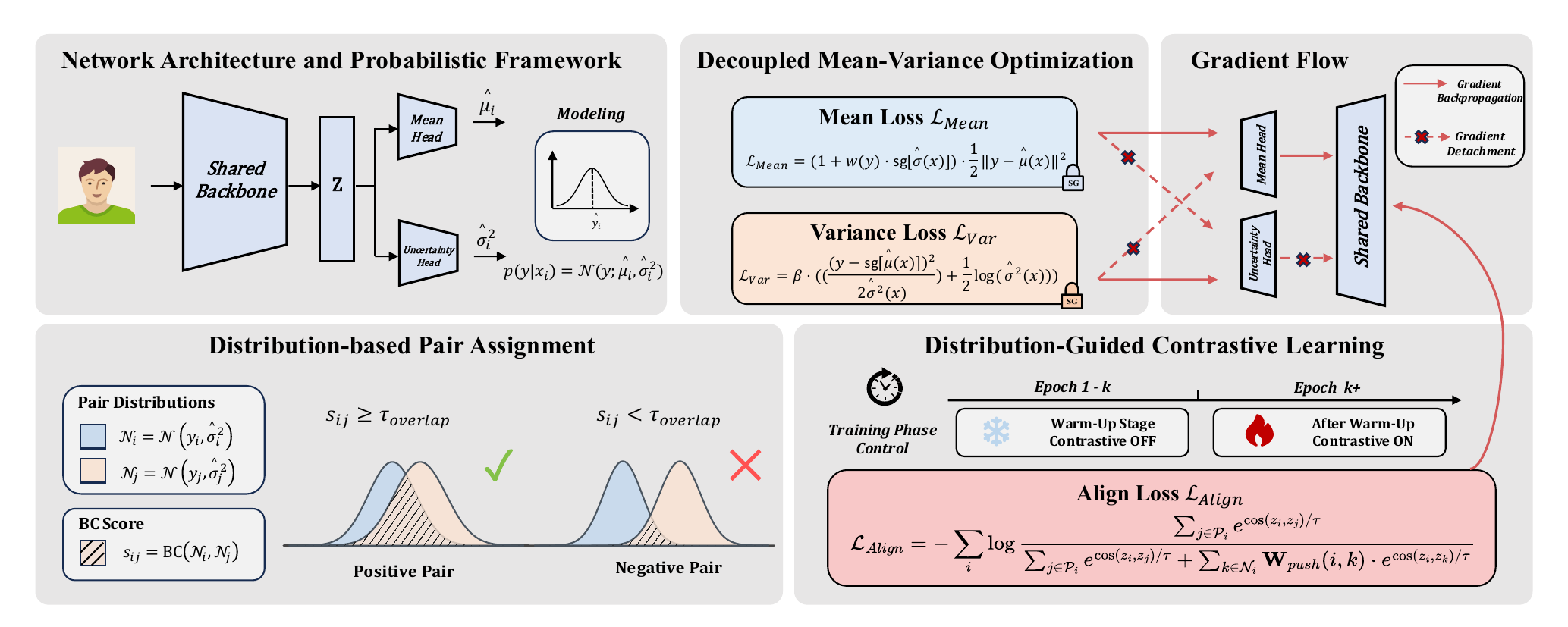}
    \caption{\textbf{Illustration of the proposed DUO framework.} Top-left: a probabilistic regression architecture that models each sample as a Gaussian distribution with predicted mean and uncertainty. Top-middle and top-right: a decoupled mean-variance optimization strategy with stop-gradient, which separates mean learning from variance fitting and explicitly controls gradient flow. Bottom-left: distribution-based pair assignment using the Bhattacharyya coefficient to determine whether two samples form a positive or negative pair according to their distribution overlap. Bottom-right: a distribution-guided contrastive learning module, which is activated after warm-up to refine the feature space for DIR.}
    \label{fig:motivation}
\end{figure*}

\subsection{Problem Formulation}
DIR aims to learn a robust model from a long-tailed dataset $\mathcal{D}=\{(x_i, y_i)\}_{i=1}^N$, where $p(y_{\text{head}}) \gg p(y_{\text{tail}})$. Conventional methods learn a deterministic mapping $f: \mathcal{X} \to \mathbb{R}$ by minimizing the Mean Squared Error (MSE). Statistically, MSE models solely the first moment (mean) of $p(y|x)$, implicitly assuming homoscedasticity (\textit{i.e.}, constant variance across all samples).

However, DIR data is inherently heteroscedastic: tail samples exhibit significantly higher uncertainty and noise compared to well-represented head samples. Ignoring this variance prevents the model from dynamically adapting to sample difficulty, fundamentally limiting its generalization on the tail.

\subsection{Probabilistic Regression Framework}

To overcome the limitations of  deterministic mappings, we reformulate the regression task as modeling the conditional probability distribution $p(y|x)$\cite{bishop1994mixture}.  Specifically, we draw inspiration from classical classification, where learning a conditional probability distribution is the standard paradigm.

\textbf{Probabilistic Modeling for Continuous Space}

In standard classification, the model essentially learns a discrete conditional probability distribution. Typically, it outputs a normalized probability vector (\textit{i.e.}, via Softmax) and makes decisions based on the Maximum A Posteriori (MAP) principle \cite{bishop2006pattern}:
\begin{equation}
\hat{y} = \arg\max_k p(y=k|x)
\end{equation}
This indicates that the core of classification lies in identifying the category with the highest confidence. From an isomorphic standpoint, regression should not be confined to predicting a single scalar; rather, it should be conceptualized as locating the point of maximum probability density within a continuous label space. To this end, we model the target conditional distribution as a Gaussian:
\begin{equation}
p(y|x; \theta) = \mathcal{N}\big(y; \mu(x), \sigma^2(x)\big)
\end{equation}

The choice of a Gaussian formulation is fundamentally principled. From an information-theoretic perspective, given the constraints of the first moment (mean) and second moment (variance), the Gaussian distribution uniquely maximizes differential entropy\cite{jaynes1957information}, thereby introducing the minimum inductive bias into our model (we provide the rigorous mathematical proof of this maximum entropy optimality in Appendix A). Under this framework, the prediction $\hat{y}$ corresponds to the mode (which is also the mean $\mu(x)$) of the distribution, mathematically unifying classification and regression.

\textbf{Instance-Level Heteroscedasticity Modeling}

The necessity of explicitly modeling the second moment $\sigma^2(x)$ stems from correcting the bias inherent in traditional methods. Although standard MSE regression is equivalent to maximizing a Gaussian likelihood, its implicit homoscedasticity assumption forces all samples to share a fixed confidence, which contradicts the intrinsic heteroscedasticity of long-tailed data.

Furthermore, since aleatoric uncertainty inherently arises from the input $x$ (e.g., occlusion or noise) rather than the target label $y$, formulating $\sigma^2(x)$ at the instance level is statistically more rigorous than relying on a class-level prior. This instance-wise formulation enables the model to dynamically quantify the specific ambiguity of each sample, providing a robust statistical foundation for regression.

\subsection{Network Architecture}

To instantiate this probabilistic framework, we minimally extend standard architectures. A shared backbone $f_{\text{enc}}$ (e.g., ResNet-50\cite{he2016deep}) first extracts features $z = f_{\text{enc}}(x) \in \mathbb{R}^D$. We then employ a dual-branch structure with a mean prediction head $\mathcal{H}_\mu$ and a parallel uncertainty head $\mathcal{H}_\sigma$. A positive parameterization $g_+$ ensures a valid standard deviation:
\begin{equation}
\begin{aligned} \hat{\mu}(x) &= W_\mu z + b_\mu, \\ \hat{\sigma}(x) &= g_+(W_\sigma z + b_\sigma), \quad g_+(\cdot)>0. \end{aligned}
\end{equation}

This design seamlessly parameterizes $p(y|x) = \mathcal{N}(y; \hat{\mu}(x), \hat{\sigma}^2(x))$ with negligible computational overhead.

\subsection{Decoupled Mean-Variance Optimization}
To train the model robustly on long-tailed distributions, we first analyze the optimization failure of the standard Negative Log-Likelihood (NLL) loss. We then propose a decoupled objective using the stop-gradient operator.

\subsubsection{\textbf{Gradient Coupling in NLL Loss}}
For heteroscedastic Gaussian regression, the standard objective is the NLL loss $\mathcal{L}_{\text{NLL}}$\cite{nix1994estimating}:
\begin{equation}
\mathcal{L}_{\text{NLL}} = \frac{1}{2} \log \hat{\sigma}^2 + \frac{(y - \hat{\mu})^2}{2\hat{\sigma}^2}
\end{equation}

To investigate its behavior under long-tailed distributions, we examine the gradient of this loss function with respect to the mean prediction $\hat{\mu}$ and variance $\hat{\sigma}^2$ :
\begin{equation}
\frac{\partial \mathcal{L}_{\text{NLL}}}{\partial \hat{\mu}} = - \frac{1}{\hat{\sigma}^2} (y - \hat{\mu}), \quad \frac{\partial \mathcal{L}_{\text{NLL}}}{\partial \hat{\sigma}^2} = \frac{\hat{\sigma}^2 - (y-\hat{\mu})^2}{2\hat{\sigma}^4}
\end{equation}

Examining the gradients reveals a \textit{gradient coupling} issue: the mean gradient $\nabla_{\hat{\mu}}\mathcal{L} = -(y-\hat{\mu})/\hat{\sigma}^2$ is scaled by $1/\hat{\sigma}^2$. As proven in Appendix B, NLL optimization drives $\hat{\sigma}^2$ toward the squared residual $(y-\hat{\mu})^2$ for hard samples (under both gradient descent and Newton's method). 

At this equilibrium, the mean gradient magnitude behaves as $\|\nabla_{\hat{\mu}}\mathcal{L}\| \approx 1/|y-\hat{\mu}| \to 0$, leading to vanishing updates for large-error samples. In the DIR setting, tail samples are typically associated with larger residuals, which further amplifies this effect: samples that require the most correction receive the weakest learning signal. Consequently, the model tends to reduce the loss by increasing the predicted variance rather than improving the mean prediction.

\subsubsection{\textbf{Decoupled Objective}}
To resolve this gradient coupling, DUO employs the stop-gradient ($\text{sg}[\cdot]$) operator\cite{van2017neural,chen2021exploring} to computationally decouple the optimization of $\hat{\mu}$ and $\hat{\sigma}$.

\textbf{Mean Loss}

To address the gradient vanishing in NLL, we utilize uncertainty to amplify rather than suppress the gradient. The mean loss is defined as:
\begin{equation}
\mathcal{L}_{Mean} = \underbrace{ (1 + w(y) \cdot \text{sg}[\hat{\sigma}(x)]) }_{\gamma: \text{Dynamic Difficulty Weight}} \cdot \frac{1}{2} \| y - \hat{\mu}(x) \|^2
\end{equation}
where the continuous label space is discretized into bins using the training set, and $w(y)$ is the inverse empirical frequency of the bin containing $y$.

The gradient with respect to $\hat{\mu}$ is $\frac{\partial \mathcal{L}_{\text{Mean}}}{\partial \hat{\mu}} = - \gamma \cdot (y - \hat{\mu})$. Here, $w(y)$ captures global label scarcity, while $\hat{\sigma}(x)$ captures instance-level difficulty. The latter provides a unidirectional forward weight but, due to $\text{sg}[\cdot]$, receives no gradient from $\mathcal{L}_{\text{Mean}}$. Thus, forward numerical dependence does not reintroduce backward optimization coupling.

\textbf{Variance Loss}

\begin{figure*}[t]
    \centering
    \includegraphics[width=1\textwidth]{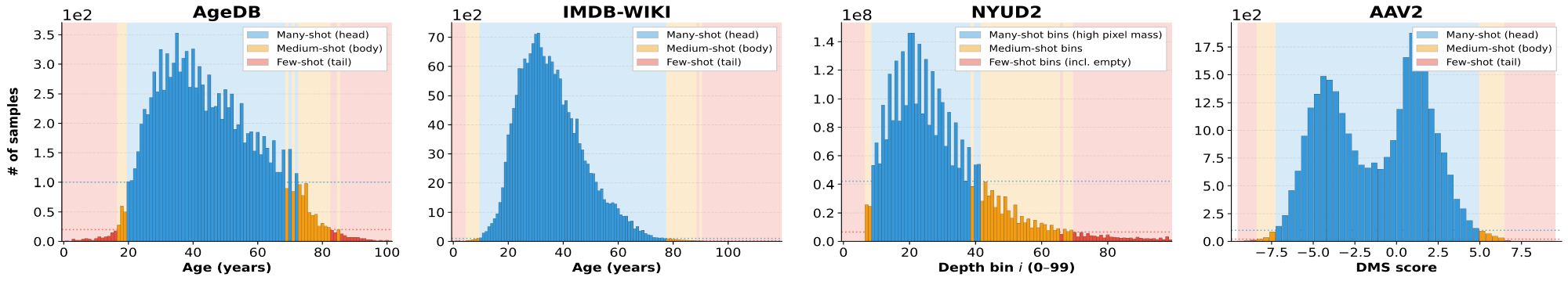}
    \caption{Overview of label distributions in the training sets for the AgeDB-DIR, IMDB-WIKI-DIR, Nydu2-DIR and AAV2-DIR datasets.}
    \label{fig:datadistribution}
\end{figure*}

Since ground-truth variance is unavailable, learning $\hat{\sigma}^2$ is unsupervised. To tightly approximate the prediction error, we design the decoupled variance loss:
\begin{equation}
\mathcal{L}_{Var} = \frac{(y - \text{sg}[\hat{\mu}(x)])^2}{2\hat{\sigma}^2(x)} + \frac{1}{2}\log\hat{\sigma}^2(x)
\end{equation}

This objective balances residual fitting and variance regularization. Specifically, the $\log \hat{\sigma}^2$ term penalizes unnecessarily large variance estimates, while the inverse-variance term penalizes under-estimated variance when the mean prediction error is large. Because $\operatorname{sg}[\cdot]$ stops gradients through $\hat{\mu}(x)$, the variance branch is optimized against a fixed residual target.

Setting the gradient with respect to $\hat{\sigma}^2$ to zero yields the stationary point $\hat{\sigma}^2(x) = (y - \hat{\mu}(x))^2$.

Appendix C further proves that this point is the unique global minimizer of $\mathcal{L}_{Var}$ in the linear variance space, and that under log-variance parameterization the optimization converges exponentially in the KL/Bregman sense. Therefore, $\hat{\sigma}$ serves as an implicit proxy for the empirical prediction residual.

\subsection{Distribution-Guided Contrastive Learning}
While the decoupled loss optimizes output-space predictions, long-tailed data still severely degrades feature representations, often leading to feature entanglement in tail classes. Standard contrastive learning uses rigid label distance thresholds (\textit{e.g.}, $|y_i - y_j| < \epsilon$) to define positive pairs, ignoring the instance-level variance. To address this, we propose a distribution-guided contrastive learning module that dynamically shapes the feature manifold based on the geometric overlap of predicted distributions.

\subsubsection{\textbf{Pair Assignment via Distribution Overlap}}

To measure semantic similarity in the feature space, we construct a label-anchored proxy distribution. 

Since ground-truth variance is unavailable, we model each sample $x_i$ as a Gaussian proxy distribution centered at its ground-truth label $y_i$ with predicted variance $\hat{\sigma}_i^2$:

\begin{equation}
p(y\mid x_i)=\mathcal{N}(y;\, y_i,\hat{\sigma}_i^2)
\end{equation}
In this way, $\hat{\sigma}_i$ captures the uncertainty range of sample $x_i$ in label space, while the ground-truth label $y_i$ provides a stable semantic anchor for pair assignment.

We quantify the geometric overlap between two proxy distributions using the Bhattacharyya Coefficient (BC)\cite{bhattacharyya1943measure,kailath1967divergence}, defined as the overlap integral of two probability density functions:
\begin{equation}
\text{BC}(i, j) = \int \sqrt{p(y|x_i) \cdot p(y|x_j)} \, dy
\end{equation}
Substituting the Gaussian proxy distributions into the overlap integral yields the following closed-form solution, whose full derivation is provided in Appendix D:
\begin{equation}
\text{BC}(i, j) = \exp \left( - \frac{1}{4} \frac{(y_i - y_j)^2}{\hat{\sigma}_i^2 + \hat{\sigma}_j^2} - \frac{1}{2} \ln \frac{\hat{\sigma}_i^2 + \hat{\sigma}_j^2}{2\hat{\sigma}_i\hat{\sigma}_j} \right) 
\end{equation}
Essentially, this metric introduces an adaptive semantic tolerance. For high-uncertainty tail samples (large $\hat{\sigma}$), BC remains high even if the label distance $|y_i - y_j|$ is large, provided the distributions overlap. This allows hard samples to interact with a broader set of semantic neighbors during manifold alignment. 

Based on this probabilistic similarity, we dynamically define the positive and negative sample sets respectively as
\begin{equation}
\begin{split}
\mathcal{P}_i &= \{j \mid \mathrm{BC}(i,j)\ge\tau_{\mathrm{overlap}}\} \\
\mathcal{N}_i &= \{j \mid \mathrm{BC}(i,j) < \tau_{\mathrm{overlap}}\}
\end{split}
\end{equation}
thereby overcoming the limitations of rigid distance-based thresholds.

\subsubsection{\textbf{Distance-Aware Contrastive Objective}}

To correct feature entanglement where samples with large label discrepancies are erroneously clustered, we design a distance-aware InfoNCE loss $\mathcal{L}_{Align}$ to geometrically constrain the backbone features $z$:

\begin{equation}
\mathcal{L}_{Align} = - \sum_{i} \log \frac{ \sum_{j \in \mathcal{P}_i} e^{\text{cos}(z_i, z_j)/\tau} }{ \sum_{j \in \mathcal{P}_i} e^{\text{cos}(z_i, z_j)/\tau} + \sum_{k \in \mathcal{N}_i} \mathbf{W}_{push}(i,k) \cdot e^{\text{cos}(z_i, z_k)/\tau} }
\end{equation}
To impose stronger repulsion on semantically incompatible yet feature-similar negatives, we introduce a dynamic penalty weight defined as the inverse Bhattacharyya overlap:

\begin{equation}
\mathbf{W}_{push}(i,k) = w(y_i) \cdot \text{BC}(i,k)^{-1}
\end{equation}
Here, $w(y_i)$ amplifies the gradient contribution of tail samples, while the inverse distribution overlap $\text{BC}(i,k)^{-1}$ heavily penalizes hard negatives with vanishing semantic overlaps. Appendix E shows that this weighting induces exponentially stronger repulsion for erroneously entangled negatives, helping restore an ordered manifold topology.

During the first $T_w$ warm-up epochs, we optimize only $\mathcal{L}_{\mathrm{Mean}} + \mathcal{L}_{\mathrm{Var}}$ to obtain stable uncertainty estimates. After this warm-up phase, $\mathcal{L}_{\mathrm{Align}}$ is activated.

\subsection{Overall Objective Function}
To jointly optimize the mean prediction, variance estimation, and feature manifold, the overall objective function is formulated as:
\begin{equation}
\mathcal{L}_{Total} = \mathcal{L}_{Mean} + \mathcal{L}_{Var} + \lambda \mathcal{L}_{Align}
\end{equation}
where $\lambda$ is a trade-off hyperparameter that balances the contribution of the alignment objective against the regression and variance estimation terms.

\section{Experiment}


We evaluate the proposed DUO on four deep imbalanced regression (DIR) datasets, spanning diverse tasks including age estimation, depth estimation, and protein mutation activity prediction. Specifically, AgeDB-DIR\cite{moschoglou2017agedb} and IMDB-WIKI-DIR\cite{rothe2018deep} are utilized for large-scale continuous facial age regression. NYUD2-DIR, built upon NYU Depth V2\cite{silberman2012indoor}, aims to predict depth maps from indoor RGB images. Furthermore, to validate the generalization capability of our method in bioinformatics, we introduce the AAV2-DIR dataset, which is constructed from high-throughput viability data of AAV2 capsid protein variants\cite{bryant2021deep}. This comprehensive experimental setup covers typical scenarios ranging from 1D continuous targets to high-dimensional structured labels, designed to thoroughly verify the effectiveness and robustness of DUO in handling various extremely imbalanced data distributions.The label distributions of the four datasets are shown in the figure \ref{fig:datadistribution}. Complete experimental results are provided in Tables S5--S8 of the Appendix.

\begin{table*}[htbp]
  \centering
  \caption{Results are presented for the few-shot region on the IMDB, AgeDB, and AAV2 datasets. The first section reports the results of the baselines and our method, with the best results highlighted in bold. The second section reports the performance differences with respect to the corresponding baselines}
  \label{tab:few_shot_results}
  \setlength{\tabcolsep}{4pt}
  \begin{tabular}{lccccccccc}
    \toprule
    & \multicolumn{3}{c}{MAE} & \multicolumn{3}{c}{bMAE} & \multicolumn{3}{c}{GM}\\
    \cmidrule(lr){2-4} \cmidrule(lr){5-7} \cmidrule(lr){8-10}
    & IMDB-DIR & AgeDB-DIR & AAV2-DIR & IMDB-DIR & AgeDB-DIR & AAV2-DIR & IMDB-DIR & AgeDB-DIR & AAV2-DIR \\
    \midrule
    Vanilla              & 25.546 & 13.381 & 7.365 & 33.085 & 16.579 & 7.565 & 18.091 & 9.644 & 7.351 \\
    LDS                  & \textbf{{22.755}} & 11.489 & 4.361 & 30.179 & 12.055 & 4.681 & 13.786 & 7.577 & 4.054 \\
    FDS                  & 24.347 & 12.136 & 5.023 & 31.839 & 15.531 & 5.388 & 14.600 & 8.326 & 4.786 \\
    Ranksim              & 25.317 & 14.601 & 4.885 & 32.464 & 17.611 & 5.225 & 17.639 & 11.021 & 4.652 \\
    ConR                 & 25.969 & 12.333 & 4.183 & 33.601 & 14.486 & 4.335 & 17.990 & 9.376 & 3.932 \\
    Balanced MSE         & 23.541 & \textbf{{9.690}} & 7.283 & 30.000 & 11.940 & 7.563 & 13.803 & 10.480 & 7.253 \\
    Distloss             & 24.215 & 12.121 & 7.432 & 30.165 & 13.670 & 7.567 & 14.076 & 7.726 & 7.416 \\
    \textbf{DUO (Ours)}  & 22.940 & 9.917 & \textbf{{4.106}} & \textbf{{27.190}} & \textbf{{11.590}} & \textbf{{4.329}} & \textbf{{13.650}} & \textbf{{6.587}} & \textbf{{3.777}} \\
    \midrule
    Ours vs. Vanilla     & {+ 2.606} & {+ 3.464} & {+ 3.259} & {+ 5.895} & {+ 4.989} & {+ 3.236} & {+ 4.441} & {+ 3.057} & {+ 3.574} \\
    Ours vs. LDS         & {- 0.185} & {+ 1.572} & {+ 0.255} & {+ 2.989} & {+ 0.465} & {+ 0.352} & {+ 0.136} & {+ 0.990} & {+ 0.277} \\
    Ours vs. FDS         & {+ 1.407} & {+ 2.219} & {+ 0.917} & {+ 4.649} & {+ 3.941} & {+ 1.059} & {+ 0.950} & {+ 1.739} & {+ 1.009} \\
    Ours vs. Ranksim     & {+ 2.377} & {+ 4.684} & {+ 0.779} & {+ 5.274} & {+ 6.021} & {+ 0.896} & {+ 3.989} & {+ 4.434} & {+ 0.875} \\
    Ours vs. ConR        & {+ 3.029} & {+ 2.416} & {+ 0.077} & {+ 6.411} & {+ 2.896} & {+ 0.006} & {+ 4.340} & {+ 2.789} & {+ 0.155} \\
    Ours vs. BM& {+ 0.601} & {- 0.227} & {+ 3.177} & {+ 2.810} & {+ 0.350} & {+ 3.234} & {+ 0.153} & {+ 3.893} & {+ 3.476} \\
    Ours vs. Distloss    & {+ 1.275} & {+ 2.204} & {+ 3.326} & {+ 2.975} & {+ 2.080} & {+ 3.238} & {+ 0.426} & {+ 1.139} & {+ 3.639} \\
    \bottomrule
  \end{tabular}
\end{table*}

\subsection{Evaluation protocol and metrics}
Following the standard DIR evaluation protocol, we report performance across four shot-based partitions: All (full test set), Many (>100 training samples per bin), Median (20–100 samples), and Few (<20 samples). For AgeDB-DIR, IMDB-WIKI-DIR and AAV2-DIR, we use MAE, balanced MAE (bMAE)~\cite{ren2022balanced}, and error geometric mean (GM)~\cite{yang2021delving}, which measure overall error, bin-balanced error, and the geometric mean of per-sample errors, respectively. For NYUD2-DIR, we follow prior depth estimation DIR work and adopt RMSE and threshold accuracy $\delta_1$ (percentage of pixels with $\max(\frac{d}{g},\frac{g}{d}) < 1.25$, $g$=ground-truth depth, $d$=predicted depth).

\subsubsection{\textbf{Main results for age estimation}}
Table \ref{tab:few_shot_results} presents results on AgeDB-DIR and IMDB-WIKI-DIR, averaged over five random runs. DUO obtains the best few-shot bMAE and GM on both benchmarks. Its few-shot MAE remains competitive but is slightly higher than LDS on IMDB-WIKI-DIR (22.940 vs. 22.755) and Balanced MSE on AgeDB-DIR (9.917 vs. 9.690), revealing an empirical trade-off between tail-balanced performance and the best metric-specific error.
This reflects an inherent head-tail trade-off in DIR, where minor head compromises (e.g., AgeDB Many-shot MAE 6.980 vs. Vanilla 6.743) are outweighed by substantial tail gains (Few-shot 9.917 vs. 13.381).

\begin{table}[htbp]
  \centering
  \caption{Experimental results on the few-shot region of the NYUD2 dataset. Note that we do not report results for two-stage training methods like Balanced MSE and DistLoss on NYUD2, as these methods are primarily designed for global regression and do not directly scale to structured pixel-wise depth prediction.}
  \label{tab:nyud2_results_complete}
  \begin{tabular}{lcccc}
    \toprule
    Method & RMSE $\downarrow$ & MAE $\downarrow$ & bMAE $\downarrow$ & $\delta_1 \uparrow$ \\
    \midrule
    Vanilla              & 1.867 & 1.443 & 1.884 & 0.608 \\
    LDS                  & 1.764 & \textbf{{1.333}} & 1.714 & 0.641 \\
    FDS                  & 1.905 & 1.478 & 1.920 & 0.594 \\
    RankSim              & 1.896 & 1.475 & 1.907 & 0.601 \\
    ConR                 & 1.869 & 1.459 & 1.912 & 0.591 \\
    \textbf{DUO(Ours)} & \textbf{{1.759}} & 1.351 & \textbf{{1.696}} & \textbf{{0.643}} \\
    \midrule
    Ours vs. Vanilla     & {+ 0.108} & {+ 0.092} & {+ 0.188} & {+ 0.035} \\
    Ours vs. LDS         & {+ 0.005} & {- 0.018} & {+ 0.018} & {+ 0.002} \\
    Ours vs. FDS         & {+ 0.146} & {+ 0.127} & {+ 0.224} & {+ 0.049} \\
    Ours vs. RankSim     & {+ 0.137} & {+ 0.124} & {+ 0.211} & {+ 0.042} \\
    Ours vs. ConR        & {+ 0.110} & {+ 0.108} & {+ 0.216} & {+ 0.052} \\
    \bottomrule
  \end{tabular}
\end{table}

\subsubsection{\textbf{Main results for depth estimation}}
We further evaluate DUO on NYUD2-DIR to examine its effectiveness in structured regression with high-dimensional depth maps. Compared with existing DIR baselines, DUO achieves the best few-shot bMAE of \textbf{1.696}, as shown in Table \ref{tab:nyud2_results_complete}. These results indicate that DUO can be effectively extended from scalar regression to more complex dense prediction settings while preserving robust performance in tail regions.

 \subsubsection{\textbf{Main results for protein activity estimation}}
Table \ref{tab:few_shot_results} also summarizes AAV2-DIR, which maps protein sequences to functional fitness. DUO achieves the best few-shot MAE, bMAE, and GM of \textbf{4.106}, \textbf{4.329}, and \textbf{3.777}, respectively. These results show that its few-shot advantages extend beyond visual regression to biological sequence activity prediction.

\subsubsection{\textbf{Feature visualizations}}

\begin{figure}[t]
    \centering
    \includegraphics[width=1\columnwidth]{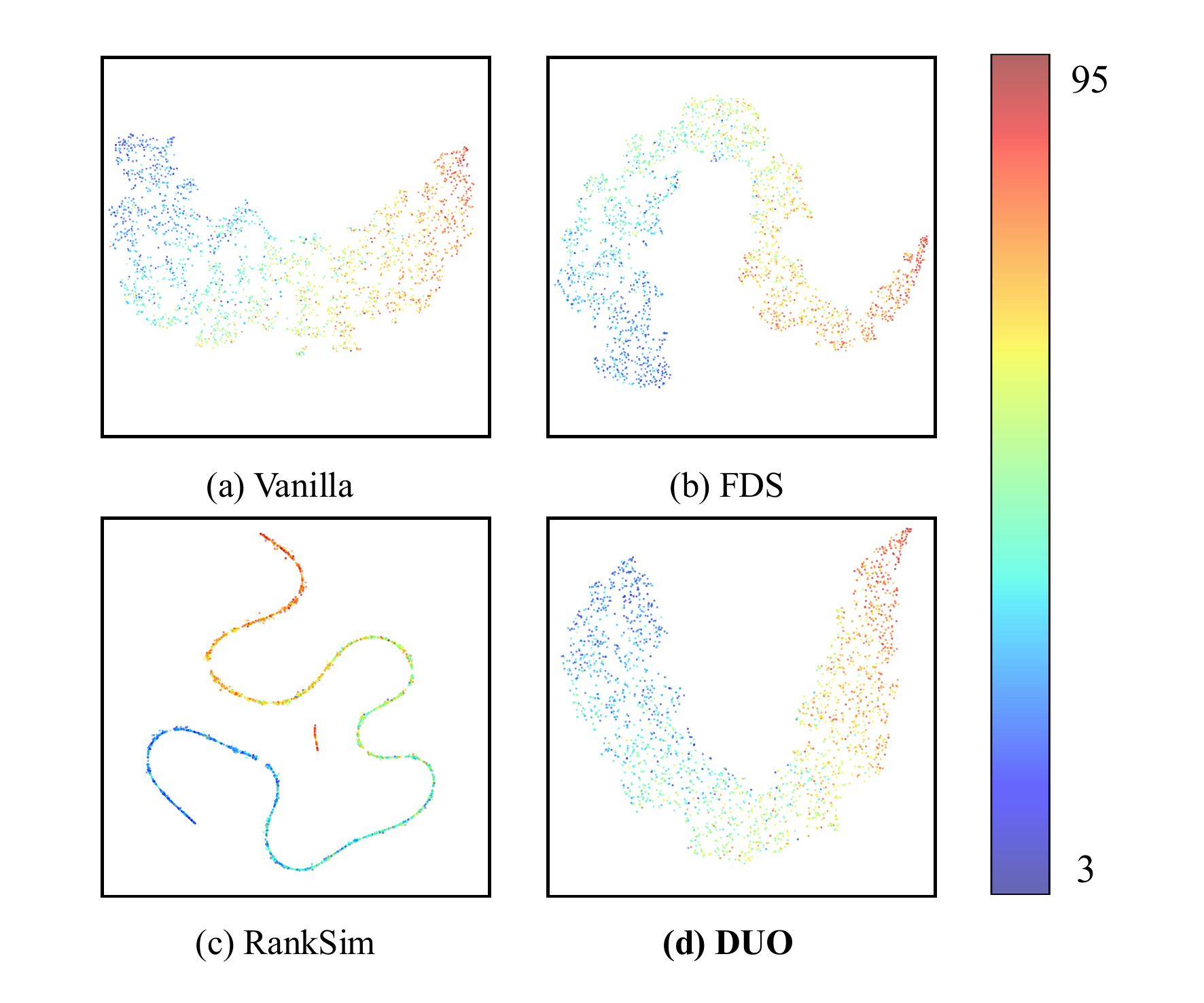}
    \caption{Feature visualization on AgeDB-DIR for (a) VANILLA, (b) FDS, (c) Ranksim and (d) \textbf{DUO}.}
    \label{fig:featurevisual}
\end{figure}

\begin{table*}[htbp]
  \centering
  \caption{Component-wise ablation study of DUO in the few-shot regions. Vanilla denotes the baseline trained with Mean Squared Error (MSE). DUO-C removes the contrastive term, and DUO uses the complete objective. The best results are highlighted in bold.}
  \label{tab:ablation_study}
  \setlength{\tabcolsep}{4.5pt}
  \begin{tabular}{lccccccccc}
    \toprule
    & \multicolumn{3}{c}{MAE $\downarrow$} & \multicolumn{3}{c}{bMAE $\downarrow$} & \multicolumn{3}{c}{GM $\downarrow$}\\
    \cmidrule(lr){2-4} \cmidrule(lr){5-7} \cmidrule(lr){8-10}
    Method & IMDB & AgeDB & AAV2 & IMDB & AgeDB & AAV2 & IMDB & AgeDB & AAV2 \\
    \midrule
    Vanilla      & 25.546 & 13.381 & 7.365 & 33.085 & 16.579 & 7.565 & 18.091 & 9.644 & 7.351 \\
    DUO-C        & 23.332 & 10.787 & 4.193 & 30.349 & 13.092 & 4.400 & 13.862 & 6.808 & 3.948 \\
    \textbf{DUO} & \textbf{{22.940}} & \textbf{{9.917}} & \textbf{{4.106}} & \textbf{{27.190}} & \textbf{{11.590}} & \textbf{{4.329}} & \textbf{{13.650}} & \textbf{{6.587}} & \textbf{{3.777}} \\
    \bottomrule
  \end{tabular}
\end{table*}

To examine representation differences on deep imbalanced regression (DIR) tasks, we use t-SNE\cite{van2008visualizing} to project the penultimate-layer features of ResNet-50 into 2D and color samples by their continuous targets.

As shown in Figure~\ref{fig:featurevisual}, Vanilla exhibits central collapse and blurred head--tail boundaries, FDS retains substantial local overlap, and RankSim forms an over-compressed line-like manifold. DUO instead produces a more continuous and less entangled layout. Since t-SNE is qualitative, Appendix Table S3 further reports KNN-MAE, where DUO obtains the lowest errors in the compound-tail and median-shot regions, quantitatively supporting improved local smoothness.

\subsubsection{\textbf{Pairwise comparison}}

Although our approach entails architectural modifications, we further evaluate its complementarity with existing methods. Table \ref{tab:pairwise_comparison_aav2} compares AAV2 baselines with their DUO-enhanced counterparts, with the better value in each pair highlighted in bold. DUO integration improves MAE and bMAE across all pairs and improves most GM values, although RankSim's GM slightly worsens. This indicates useful but metric-dependent complementarity. Results on the other datasets are provided in Appendix Tables S9--S11.

\begin{table*}[htbp]
  \centering
  \caption{Pairwise comparison of various baselines and their DUO-enhanced counterparts on the AAV2 dataset across different shot regions. The best results within each comparison pair are highlighted in bold.}
  \label{tab:pairwise_comparison_aav2}
  \resizebox{\textwidth}{!}{
  \begin{tabular}{lcccccccccccc}
    \toprule
    & \multicolumn{3}{c}{Overall} & \multicolumn{3}{c}{Few} & \multicolumn{3}{c}{Median} & \multicolumn{3}{c}{Many} \\
    \cmidrule(lr){2-4} \cmidrule(lr){5-7} \cmidrule(lr){8-10} \cmidrule(lr){11-13}
    Method & MAE $\downarrow$ & bMAE $\downarrow$ & GM $\downarrow$ & MAE $\downarrow$ & bMAE $\downarrow$ & GM $\downarrow$ & MAE $\downarrow$ & bMAE $\downarrow$ & GM $\downarrow$ & MAE $\downarrow$ & bMAE $\downarrow$ & GM $\downarrow$ \\
    \midrule
    LDS                  & 1.990 & 1.981 & 1.178 & 4.361 & 4.681 & 4.054 & 3.648 & 3.888 & 3.211 & 1.935 & 2.117 & 1.140 \\
    \textbf{LDS+DUO}     & \textbf{{1.721}} & \textbf{{1.718}} & \textbf{{1.060}} & \textbf{{4.159}} & \textbf{{4.329}} & \textbf{{3.770}} & \textbf{{3.253}} & \textbf{{3.440}} & \textbf{{2.800}} & \textbf{{1.669}} & \textbf{{1.817}} & \textbf{{1.027}} \\
    \midrule
    FDS                  & 1.988 & 1.985 & 1.256 & 5.023 & 5.388 & 4.786 & 4.345 & 4.667 & 4.008 & 1.911 & 2.147 & 1.210 \\
    \textbf{FDS+DUO}     & \textbf{{1.794}} & \textbf{{1.791}} & \textbf{{1.128}} & \textbf{{4.379}} & \textbf{{4.542}} & \textbf{{3.984}} & \textbf{{3.603}} & \textbf{{3.854}} & \textbf{{3.273}} & \textbf{{1.734}} & \textbf{{1.923}} & \textbf{{1.090}} \\
    \midrule
    RankSim              & 1.936 & 1.932 & \textbf{{1.189}} & 4.885 & 5.225 & 4.652 & 4.151 & 4.408 & 3.857 & 1.864 & 2.110 & \textbf{{1.145}} \\
    \textbf{RankSim+DUO} & \textbf{{1.831}} & \textbf{{1.829}} & 1.231 & \textbf{{4.796}} & \textbf{{5.195}} & \textbf{{4.408}} & \textbf{{4.002}} & \textbf{{4.279}} & \textbf{{3.651}} & \textbf{{1.759}} & \textbf{{1.977}} & 1.189 \\
    \midrule
    Balanced MSE (GAI)        & 2.650 & 2.639 & 2.115 & 7.283 & 7.563 & 7.253 & 5.992 & 6.226 & 5.978 & 2.539 & 2.745 & 2.045 \\
    \textbf{Balanced MSE+DUO} & \textbf{{2.439}} & \textbf{{2.441}} & \textbf{{1.839}} & \textbf{{6.677}} & \textbf{{7.076}} & \textbf{{6.576}} & \textbf{{5.693}} & \textbf{{5.968}} & \textbf{{5.598}} & \textbf{{2.333}} & \textbf{{2.573}} & \textbf{{1.775}} \\
    \bottomrule
  \end{tabular}
  }
\end{table*}

\subsection{Ablation Studies on Design Modules}

\subsubsection{\textbf{Component-wise ablation of DUO}}
Table \ref{tab:ablation_study} shows that the decoupled objective already improves few-shot performance, while the full DUO further benefits from contrastive alignment. To isolate capacity, decoupling, uncertainty weighting, and BC-based assignment, Appendix Table S1 reports controlled variants under matched dual-head capacity. The results show that capacity alone is insufficient and that removing weighting or replacing BC consistently degrades few-shot metrics.

\subsubsection{\textbf{Analysis of the hyper-parameter $ \lambda$}}

\begin{figure}[t]
    \centering
    \includegraphics[width=0.9\columnwidth]{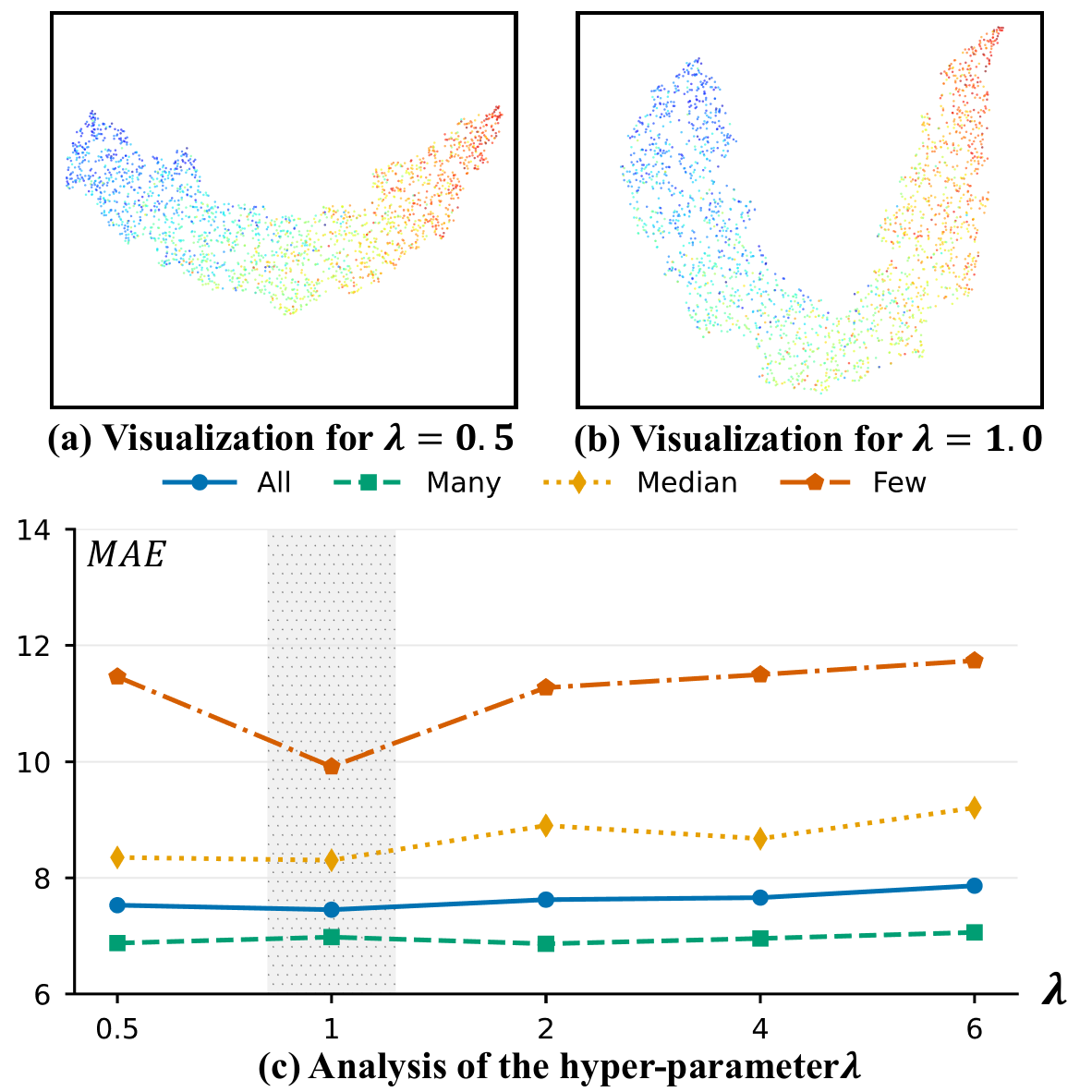}
    \caption{\textbf{Ablation study on the hyper-parameter $\lambda$. } (a) and (b) compares the learnt feature space for parameter of 0.5 and 1 respectively. (c) shows the MAE performance under various $\lambda$ values. }
    \label{fig:ablation}
\end{figure}

We further analyze the impact of hyper-parameter $\lambda$, which controls the weight of DUO’s contrastive term and governs the trade-off between sample-level fitting and structure-level constraints during optimization.

As shown in Figure \ref{fig:ablation}, a small $\lambda$ imposes insufficient contrastive constraints, resulting in training dynamics close to standard regression and limited performance gains in the few-shot region. With a moderate increase in $\lambda$, few-shot region performance improves consistently, as stronger structural constraints effectively mitigate the representation shift induced by imbalanced data distributions. However, an excessively large $\lambda$ degrades overall performance across all regions, as overly strong auxiliary constraints disrupt the primary regression objective and hinder the model from balancing performance across target intervals. Feature visualizations further corroborate these trends: a moderate $\lambda$ yields a clearer, more continuous feature structure with well-defined inter-region boundaries and minimal overlap between minority and majority samples, while an overly large $\lambda$ causes feature space over-separation or local instability. Based on both quantitative results and visual analysis, we fix $\lambda$ = 1.0 for all subsequent experiments.

\section{Conclusion}

We presented DUO, an uncertainty-aware framework for deep imbalanced regression. By modeling each target as a conditional Gaussian distribution, DUO computationally decouples mean and variance optimization and refines feature learning through distribution-guided contrastive alignment. Experiments on AgeDB-DIR, IMDB-WIKI-DIR, NYUD2-DIR, and AAV2-DIR show the strongest gains in scarce regions, particularly on imbalance-aware metrics, with modest trade-offs on some overall or many-shot results. These findings support sample-wise uncertainty as an effective signal for robust long-tailed regression.

\clearpage
\bibliographystyle{ACM-Reference-Format}
\balance
\bibliography{sample-base}


\end{document}


\title{Supplementary Material for ``Beyond Homoscedasticity: Decoupled
Uncertainty Optimization for Deep Imbalanced Regression''}
\thanks{Accepted at the 34th ACM International Conference on Multimedia
(ACM MM 2026). The Version of Record will be available at
\url{https://doi.org/10.1145/3767308.3835898}.}

\author{Juncheng Zhou}
\authornote{Juncheng Zhou and Jiaxi Lu contributed equally to this work.}
\orcid{0009-0002-2151-9058}
\affiliation{%
  \institution{School of Cyber Science and Engineering, Wuhan University}
  \city{Wuhan}
  \country{China}}
\email{2020302181088@whu.edu.cn}

\author{Jiaxi Lu}
\authornotemark[1]
\orcid{0009-0000-6123-9961}
\affiliation{%
  \institution{School of Cyber Science and Engineering, Wuhan University}
  \city{Wuhan}
  \country{China}}
\email{lulululu@whu.edu.cn}

\author{Weijing Zeng}
\orcid{0009-0009-2463-7675}
\affiliation{%
  \institution{School of Mathematics and Statistics,
Wuhan University}
  \city{Wuhan}
  \country{China}}
\email{2021302011066@whu.edu.cn}

\author{Zhong Li}
\orcid{0009-0007-0523-3638}
\affiliation{%
  \institution{School of Synthetic Biology and
Biomanufacturing, Tianjin University}
  \city{Tianjin}
  \country{China}}
\email{lz95\_@tju.edu.cn}

\author{Hao Qi}
\orcid{0000-0002-2849-6226}
\affiliation{%
  \institution{School of Synthetic Biology and
Biomanufacturing, Tianjin University}
  \city{Tianjin}
  \country{China}}
\email{haoq@tju.edu.cn}

\author{Jingsong Cui}
\correspondingauthor
\orcid{0000-0002-3111-9798}
\affiliation{%
  \institution{School of Cyber Science and Engineering, Wuhan University}
  \city{Wuhan}
  \country{China}}
\email{jscui@whu.edu.cn}

\renewcommand{\shortauthors}{Zhou et al.}
\maketitle


\appendix

\section{Proof of Maximum Entropy Optimality for Gaussian Distribution}
\label{sec:appendix_a}

In this section, we demonstrate from an information-theoretic perspective that the Gaussian distribution is the unique optimal probabilistic form that introduces the minimum inductive bias under given mean $\mu$ and variance $\sigma^2$ constraints, achieved by maximizing the differential entropy\cite{jaynes1957information}.

\begin{theorem}[Maximum Entropy Distribution]
    \label{thm:max_entropy}
    Let $X$ be a continuous random variable defined on $(-\infty, +\infty)$ with a probability density function $p(x)$. Among all possible distributions satisfying $\mathbb{E}[X] = \mu$ and $\text{Var}(X) = \sigma^2$, the unique distribution that maximizes the differential entropy $H(p)$ is the Gaussian distribution $\mathcal{N}(\mu, \sigma^2)$.
\end{theorem}

\begin{proof}
    The differential entropy of a continuous random variable $X$ is defined as $H(p) = -\int_{-\infty}^{+\infty} p(x) \ln p(x) \, dx$. To find the distribution $p(x)$ that maximizes $H(p)$, we introduce the following constraints:
    \begin{itemize}
        \item Normalization: $\int_{-\infty}^{+\infty} p(x) \, dx = 1$
        \item Mean: $\int_{-\infty}^{+\infty} x p(x) \, dx = \mu$
        \item Variance: $\int_{-\infty}^{+\infty} (x-\mu)^2 p(x) \, dx = \sigma^2$
    \end{itemize}
    We construct the Lagrangian functional $\mathcal{L}(p)$ incorporating these constraints:
    \begin{align*}
        \mathcal{L}(p) &= - \int_{-\infty}^{+\infty} p(y) \ln p(y) \, dy + \lambda_0 \left( \int_{-\infty}^{+\infty} p(y) \, dy - 1 \right) \\
        &\quad + \lambda_1 \left( \int_{-\infty}^{+\infty} y p(y) \, dy - \mu \right) + \lambda_2 \left( \int_{-\infty}^{+\infty} (y - \mu)^2 p(y) \, dy - \sigma^2 \right)
    \end{align*}
    Setting the functional derivative with respect to $p(y)$ to zero yields the extremum condition:
    \begin{equation*}
        \frac{\delta \mathcal{L}}{\delta p(y)} = -1 - \ln p(y) + \lambda_0 + \lambda_1 y + \lambda_2 (y - \mu)^2 = 0
    \end{equation*}
    which implies $p(y) = \exp\left( (\lambda_0 - 1) + \lambda_1 y + \lambda_2 (y - \mu)^2 \right)$. 
    For integrability we require $\lambda_2 < 0$.
    And for a Gaussian-type distribution $p(y) \propto e^{a y^2 + b y}$ with $a<0$, the mean is
    \begin{equation*}
        \mathbb{E}[y] = -\frac{b}{2a}.
    \end{equation*}
    Substituting $a = \lambda_2$ and $b = \lambda_1 - 2\lambda_2 \mu$ yields
    \[
    \mathbb{E}[y] = -\frac{\lambda_1 - 2\lambda_2 \mu}{2\lambda_2} = \mu - \frac{\lambda_1}{2\lambda_2}.
    \]
    Using the mean constraint $\mathbb{E}[y] = \mu$, we have
    \[
    \mu - \frac{\lambda_1}{2\lambda_2} = \mu.
    \]
    Since $\lambda_2 < 0$, we know $2\lambda_2 \neq 0$, which implies
    \[
    \lambda_1 = 0.
    \]  
    Let $C = \exp(\lambda_0 - 1)$. Substituting $p(y)$ into the constraints and utilizing Gaussian integrals:
    \begin{align*}
        1 &= \int_{-\infty}^{+\infty} C e^{\lambda_2 (y - \mu)^2} \, dy = C \sqrt{\frac{\pi}{-\lambda_2}}, \quad \sigma^2 = \int_{-\infty}^{+\infty} (y - \mu)^2 C e^{\lambda_2 (y - \mu)^2} \, dy = \frac{C\sqrt{\pi}}{2(-\lambda_2)^{3/2}} \\
        \implies \sigma^2 &= \frac{C\sqrt{\pi} / 2(-\lambda_2)^{3/2}}{C\sqrt{\pi / -\lambda_2}} = \frac{1}{-2\lambda_2} \implies \lambda_2 = -\frac{1}{2\sigma^2}, \quad C = \frac{1}{\sqrt{2\pi\sigma^2}}
    \end{align*}
    Substituting $\lambda_2$ and $C$ back into the probability density function, we obtain:
    \begin{equation}
        p(y) = \frac{1}{\sqrt{2\pi\sigma^2}} \exp \left( - \frac{(y - \mu)^2}{2\sigma^2} \right)
    \end{equation}
    The derivation rigorously establishes the optimal probabilistic form for instance-level variance modeling from an information-theoretic perspective.
\end{proof}

\section{Optimization Laziness Analysis of NLL Loss in Deep Imbalanced Regression}
\label{sec:appendix_b}
\label{app:nll_analysis}

\subsection{Problem Formulation and Notation}
Consider a heteroscedastic Gaussian regression model where for an input $x$, the model outputs mean $\mu(x)$ and variance $\sigma^2(x) > 0$. Given label $y$, the single-sample Negative Log-Likelihood (NLL) loss is \cite{nix1994estimating}:
\begin{equation}
    \mathcal{L}_{NLL} = \frac{1}{2}\log\sigma^2(x) + \frac{(y - \mu(x))^2}{2\sigma^2(x)}
\end{equation}
For simplicity, we define the residual $e(x) := y - \mu(x)$ and let $v := \sigma^2(x) > 0$ be the optimization variable. The loss becomes $\mathcal{L} = \frac{1}{2}\log v + \frac{e^2}{2v}$. In this analysis, we study partial derivatives with respect to $\mu$ and $v$ separately; hence, when differentiating with respect to $v$, the residual $e=y-\mu$ is treated as fixed.

\subsection{Gradient Computation}
\begin{proposition}
    \label{ap:b1}
    The first-order partial derivatives of $\mathcal{L}$ with respect to $\mu$ and $v = \sigma^2$ are:
    \begin{equation}
        \nabla_{\mu}\mathcal{L} = \frac{\mu - y}{\sigma^2(x)} = -\frac{e}{\sigma^2}, \quad \nabla_{v}\mathcal{L} = \frac{\sigma^2(x) - e^2}{2\sigma^4(x)}
    \end{equation}
\end{proposition}

\begin{proof}
    Fixing $v$, the derivative w.r.t. $\mu$ is:
    \begin{equation*}
        \frac{\partial \mathcal{L}}{\partial \mu} = \frac{\partial}{\partial \mu}\left[\frac{(y-\mu)^2}{2v}\right] = \frac{2(y-\mu)(-1)}{2v} = \frac{\mu - y}{v} = -\frac{e}{\sigma^2}
    \end{equation*}
    where $\frac{\partial e}{\partial \mu} = -1$. The mean gradient is scaled by $1/\sigma^2(x)$; thus, a large $\sigma^2(x)$ directly suppresses the update step of $\mu$. 
    
    Fixing $\mu$, the derivative w.r.t. $v$ is:
    \begin{equation*}
        \frac{\partial \mathcal{L}}{\partial v} = \frac{\partial}{\partial v}\left[\frac{1}{2}\log v + \frac{e^2}{2v}\right] = \frac{1}{2v} - \frac{e^2}{2v^2} = \frac{v - e^2}{2v^2} = \frac{\sigma^2(x) - e^2}{2\sigma^4(x)}
    \end{equation*}
\end{proof}

\subsection{Convergence of Standard Iterative Methods to a Unique Optimum}

\subsubsection{Gradient Descent (GD)}
\begin{theorem}[Variance Inflation and Convergence]
    Under the GD update rule with learning rate $\eta > 0$, if $e^2 > v_t$, the variance strictly increases. The unique stable equilibrium is $v^* = e^2$, around which GD converges linearly.
\end{theorem}

\begin{proof}
    \textit{(a) Monotonicity:} The GD rule is $v_{t+1} = v_t - \eta \frac{\partial \mathcal{L}}{\partial v}\big|_{v_t} = v_t - \eta \frac{v_t - e^2}{2v_t^2}$. If $e^2 > v_t$, then $\frac{\partial \mathcal{L}}{\partial v} < 0$, leading to $v_{t+1} > v_t$. Thus, the variance strictly increases.

    \textit{(b) Uniqueness, Lipschitz Smoothness and Step Size Condition:}
    Setting $\frac{\partial \mathcal{L}}{\partial v} = 0$ yields the unique critical point $v^* = e^2$.
    The second-order derivative (Hessian) is
    \[
    \frac{\partial^2 \mathcal{L}}{\partial v^2} = \frac{\partial}{\partial v} \left( \frac{1}{2v} - \frac{e^2}{2v^2} \right) = -\frac{1}{2v^2} + \frac{e^2}{v^3} = \frac{2e^2 - v}{2v^3}.
    \]
    In a neighborhood of $v^* = e^2$, the Hessian is positive and bounded above by $L = \frac{1}{2e^4}$, which implies the gradient of $\mathcal{L}$ is $L$-Lipschitz continuous in this neighborhood.
    For gradient descent to converge stably to a local minimum, the step size must satisfy $0 < \eta < \frac{2}{L} = 4e^4$.
    At $v = e^2$, we have $\left. \frac{\partial^2 \mathcal{L}}{\partial v^2} \right|_{v=e^2} = \frac{1}{2e^4} > 0$, so $v^* = e^2$ is a strict local minimum.

    \textit{(c) Linear Convergence Rate:} Let $\delta_t = v_t - e^2$. A first-order Taylor expansion near $v^*$ gives $\delta_{t+1} \approx (1 - \frac{\eta}{e^4})\delta_t + \mathcal{O}(\delta_t^2)$. With $0 < \eta < 2e^4$, the convergence factor $\rho = |1 - \eta/e^4| < 1$ ensures linear convergence: $\delta_t = \mathcal{O}(\rho^t)$.
\end{proof}
In summary, under the approximate equilibrium state $\sigma^2(x) \approx e(x)^2$, the predicted variance of "hard" samples (where $e^2 \gg \sigma^2$) will rapidly inflate to the magnitude of the squared residual.

\subsubsection{Newton's Method}

\begin{theorem}[Quadratic Convergence under Newton's Method]
    \label{thm:newton_convergence}
    When updating the variance using Newton's method, the iteration satisfies:
    \begin{equation}
        v_{t+1} = \frac{v_t(3e^2 - 2v_t)}{2e^2 - v_t}
    \end{equation}
    Furthermore, the deviation $\delta_t = v_t - e^2$ achieves quadratic convergence:
    \begin{equation}
        |\delta_{t+1}| = \left|\frac{2\delta_t^2}{e^2 - \delta_t}\right| = \mathcal{O}(\delta_t^2)
    \end{equation}
\end{theorem}

\begin{proof}
    (a) \textit{Derivation of Newton's update formula:} The update rule is $v_{t+1} = v_t - [\nabla^2_v \mathcal{L}]^{-1} \nabla_v \mathcal{L}$. Substituting the known derivatives $\nabla_v \mathcal{L} = \frac{v - e^2}{2v^2}$ and $\nabla^2_v \mathcal{L} = \frac{2e^2 - v}{2v^3}$, the iteration step size $\Delta v$ and the updated variance $v_{t+1}$ are computed as:
    \begin{align*}
        \Delta v &= -\frac{2v^3}{2e^2 - v} \cdot \frac{v - e^2}{2v^2} = \frac{v(e^2 - v)}{2e^2 - v} \\
        v_{t+1} &= v + \frac{v(e^2 - v)}{2e^2 - v} = \frac{v(2e^2 - v) + v(e^2 - v)}{2e^2 - v} = \frac{v(3e^2 - 2v)}{2e^2 - v}
    \end{align*}

    (b) \textit{Verification of the equilibrium point:} Letting $v = e^2$, we obtain $v_{t+1} = \frac{e^2(3e^2 - 2e^2)}{2e^2 - e^2} = e^2$.

    (c) \textit{Quadratic convergence rate:} Newton's method converges quadratically to $v^*$ for initial points sufficiently close to $v^*$, since the Hessian $\nabla_v^2 \mathcal{L}(v^*) = \frac{1}{2e^4} > 0$ is positive definite and non-singular at $v^*$. Let $\delta_t = v_t - e^2$. Substituting $v_t = e^2 + \delta_t$ into the update formula, the deviation at the next step $\delta_{t+1} = v_{t+1} - e^2$ is:
    \begin{align*}
        \delta_{t+1} &= \frac{(e^2 + \delta_t)(e^2 - 2\delta_t)}{e^2 - \delta_t} - e^2 = \frac{e^4 - e^2\delta_t - 2\delta_t^2 - e^4 + e^2\delta_t}{e^2 - \delta_t} = \frac{-2\delta_t^2}{e^2 - \delta_t}
    \end{align*}
    For sufficiently small $\delta_t$, $e^2 - \delta_t > 0$, ensuring $v_t > 0$ throughout the iteration.
    Thus, $|\delta_{t+1}| = \frac{2\delta_t^2}{|e^2 - \delta_t|} = \mathcal{O}(\delta_t^2)$. This demonstrates that the residual converges quadratically.
\end{proof}

\subsection{Proof of the Vanishing Mean Update Signal}
\label{sec:vanishing_mean}

\begin{theorem}[Gradient Vanishing]
    \label{thm:gradient_vanishing}
    Under the approximate equilibrium state $\sigma^2(x) \approx e(x)^2$:
    \begin{equation}
        \left\|\nabla_{\mu}\mathcal{L}\right\| = \frac{|e|}{\sigma^2} \approx \frac{|e|}{e^2} = \frac{1}{|e|}
    \end{equation}
    Consequently, $\lim_{|e(x)| \to \infty} \left\|\nabla_{\mu}\mathcal{L}\right\| = 0$. That is, the mean update signal for hard samples (with extremely large errors) tends to vanish.
\end{theorem}

\begin{proof}
    we have $\nabla_\mu \mathcal{L} = -\frac{e}{\sigma^2}$, thus the gradient magnitude is:
    \[
    \|\nabla_\mu \mathcal{L}\| = \frac{|e|}{\sigma^2}.
    \]
    By the equilibrium condition, $\sigma^2 = e^2 + o(e^2)$ as $|e| \to \infty$, which means $\lim_{|e| \to \infty} \frac{\sigma^2 - e^2}{e^2} = 0$. Substituting into the gradient:
    \[
    \frac{|e|}{\sigma^2} = \frac{|e|}{e^2 + o(e^2)} = \frac{1}{|e|} \cdot \frac{1}{1 + \frac{o(e^2)}{e^2}}.
    \]
    As $|e| \to \infty$, $\frac{o(e^2)}{e^2} \to 0$, so $\frac{1}{1 + \frac{o(e^2)}{e^2}} \to 1$, hence:
    \[
    \frac{|e|}{\sigma^2} \sim \frac{1}{|e|} \quad (|e| \to \infty).
    \]
    Taking the limit:
    \[
    \lim_{|e| \to \infty} \|\nabla_\mu \mathcal{L}\| = \lim_{|e| \to \infty} \frac{1}{|e|} = 0.
    \]
    This completes the proof that the mean update signal vanishes for hard samples.
\qed
\end{proof}

To rapidly decrease the NLL loss for hard samples, the optimizer inflates the predicted variance. However, this simultaneously suppresses the learning signal for the mean to zero, rendering the model almost unable to improve its mean predictions for these challenging samples. 

Whether using Gradient Descent or Newton's method, the optimization of the NLL loss drives $\sigma^2(x) \to e(x)^2 = \bigl(y - \mu(x)\bigr)^2$. Since Gradient Descent converges linearly and Newton's method converges quadratically, the adaptive variance actively ``absorbs'' the errors of hard samples during training. As a result, the mean head nearly stops learning for these instances.

\section{Convergence Measure Proof for $\mathcal{L}_{Var}$ Loss}
\label{sec:appendix_c}

\textbf{Proof Objective:} For nonzero residuals, we identify the unique global minimum of $\mathcal{L}_{Var}$ and characterize its local optimization behavior in the linear and logarithmic parameterizations.

\subsection{Problem Formulation}

\subsubsection{Loss Function}
The variance-specific loss $\mathcal{L}_{Var}$ is defined as:
\begin{equation}
    \mathcal{L}_{Var} = \beta \cdot \left(\frac{(y - \operatorname{sg}[\hat{\mu}(x)])^2}{2\hat{\sigma}^2(x)} + \frac{1}{2}\log\hat{\sigma}^2(x)\right)
\end{equation}
where:
\begin{itemize}
    \item $\operatorname{sg}[\cdot]$ denotes the stop-gradient operator, which treats the mean prediction $\hat{\mu}(x)$ as a constant during variance optimization to decouple the two processes.
    \item $\beta > 0$ is a positive scaling coefficient.
    \item $\hat{\sigma}^2(x) > 0$ is the predicted variance to be optimized.
\end{itemize}

\subsubsection{Notation Simplification}
For a fixed sample $(x, y)$ with a nonzero residual, let $E := (y - \operatorname{sg}[\hat{\mu}(x)])^2 > 0$ and $S := \hat{\sigma}^2(x) > 0$. The objective function simplifies to a univariate function of $S$:
\begin{equation}
    J(S) = \beta\left(\frac{E}{2S} + \frac{1}{2}\log S\right), \quad S > 0
\end{equation}
Under logarithmic parameterization (let $v = \log \hat{\sigma}^2(x)$, hence $S = e^v$):
\begin{equation}
    \tilde{J}(v) = \frac{\beta}{2}\left(E e^{-v} + v\right), \quad v \in \mathbb{R}
\end{equation}

\subsection{Convergence Analysis in Linear Space}

\subsubsection{Global Unique Minimum}
\begin{theorem}
    The function $J(S)$ has a unique global minimum on $(0, +\infty)$ at:
    \begin{equation}
        S^* = E = (y - \hat{\mu}(x))^2
    \end{equation}
\end{theorem}

\begin{proof}
    The derivative is $\frac{\partial J(S)}{\partial S} = \frac{\beta(S-E)}{2S^2}$. It is negative for $0<S<E$, zero only at $S=E$, and positive for $S>E$. Hence, $J$ decreases up to $E$ and increases thereafter, so $S^*=E$ is the unique global minimum. Notice that $J''(S)=\frac{\beta(2E-S)}{2S^3}$ changes sign; global convexity is therefore neither claimed nor required. Thus, $\mathcal{L}_{Var}$ and $\mathcal{L}_{NLL}$ share the same stationary variance for a fixed mean prediction.
\end{proof}

\subsubsection{Convergence Rate in Linear Space}
A critical distinction lies in the gradient dynamics. For $\mathcal{L}_{NLL}$, the mean gradient $\frac{\partial \mathcal{L}_{NLL}}{\partial \mu} = -\frac{y - \mu(x)}{\sigma^2(x)}$ is suppressed by a factor of $c$ when $\sigma^2(x)$ inflates ($\sigma^2 = cE, c \gg 1$). In contrast, for $\mathcal{L}_{Var}$, the stop-gradient operator ensures $\frac{\partial \mathcal{L}_{Var}}{\partial \mu}\big|_{Var \text{ term}} \equiv 0$. The mean head is updated via an independent loss term, making its gradient magnitude invariant to $\sigma^2$ inflation.

Comparing the Hessians at $S=E$: $H_{Var} = \frac{\beta}{2E^2} = \beta \cdot H_{NLL}$. Under Gradient Descent (GD) with step size $\eta$, the spectral radii are:
\begin{equation}
    \rho_{NLL} = |1 - \frac{\eta}{2E^2}|, \quad \rho_{Var} = |1 - \frac{\eta\beta}{2E^2}|
\end{equation}
The coefficient $\beta$ therefore rescales the local curvature and must be considered jointly with the optimization step size.

\begin{theorem}[Quadratic Convergence of Newton's Method]
    Let $\delta_t = S_t - E$ be the deviation at step $t$. When initialized sufficiently close to $E$, Newton's method on $J(S)$ achieves quadratic convergence: $|\delta_{t+1}| = \mathcal{O}(\delta_t^2)$.
\end{theorem}

\begin{proof}
    The Newton update $S_{t+1} = S_t - \frac{J'(S_t)}{J''(S_t)}$ simplifies to $S_{t+1} = \frac{S_t(3E - 2S_t)}{2E - S_t}$. Substituting $S_t = E + \delta_t$:
    \begin{equation*}
        \delta_{t+1} = \frac{(E+\delta_t)(E - 2\delta_t)}{E - \delta_t} - E = \frac{E^2 - E\delta_t - 2\delta_t^2 - E^2 + E\delta_t}{E - \delta_t} = \frac{-2\delta_t^2}{E - \delta_t}
    \end{equation*}
    Thus, $|\delta_{t+1}| = \frac{2\delta_t^2}{|E - \delta_t|} = \mathcal{O}(\delta_t^2)$.
\end{proof}

\subsubsection{Convergence under $\chi^2$-Divergence}
\begin{theorem}
    Locally, GD optimization of $J(S)$ induces a contraction in the $\chi^2$-divergence space. Defining $D_{\chi^2}(S \| E) := \frac{(S - E)^2}{E}$, the iteration satisfies:
    \begin{equation}
        D_{\chi^2}(S_{t+1} \| E) = (1 - \frac{\eta \beta}{2E^2})^2 D_{\chi^2}(S_t \| E) + \mathcal{O}(|\delta_t|^3)
    \end{equation}
\end{theorem}

\begin{proof}
    Expanding $\delta_{t+1} = \delta_t - \eta \frac{\beta \delta_t}{2S_t^2}$ near $S_t \approx E$, we get $\delta_{t+1} = \rho \delta_t + \mathcal{O}(\delta_t^2)$ where $\rho := 1 - \frac{\eta\beta}{2E^2}$. For $|\rho| < 1$, squaring this local expansion gives $D_{\chi^2}(S_{t+1} \| E) = \rho^2 D_{\chi^2}(S_t \| E) + \mathcal{O}(|\delta_t|^3)$.
\end{proof}

\section{Closed-form Derivation of the Bhattacharyya Coefficient for Gaussian Distributions}
\label{sec:appendix_d}

This appendix provides the detailed mathematical derivation for the closed-form Bhattacharyya Coefficient (BC) between two label-anchored Gaussian proxy distributions, $p(y|x_i) = \mathcal{N}(y_i, \hat{\sigma}_i^2)$ and $p(y|x_j) = \mathcal{N}(y_j, \hat{\sigma}_j^2)$, as presented in Section 3.5.1. By substituting the Gaussian density functions into the overlap integral and completing the square for the exponential terms, we have:

\begin{align*} 
\text{BC}(i, j) &= \int_{-\infty}^{+\infty} \sqrt{p(y|x_i) p(y|x_j)} \, dy \\ 
&= \int_{-\infty}^{+\infty} \sqrt{ \frac{1}{\sqrt{2\pi\hat{\sigma}_i^2}} \exp\left(-\frac{(y-y_i)^2}{2\hat{\sigma}_i^2}\right) \cdot \frac{1}{\sqrt{2\pi\hat{\sigma}_j^2}} \exp\left(-\frac{(y-y_j)^2}{2\hat{\sigma}_j^2}\right) } \, dy \\ 
&= \frac{1}{(2\pi\hat{\sigma}_i\hat{\sigma}_j)^{1/2}} \int_{-\infty}^{+\infty} \exp\left( -\frac{1}{4} \left[ \frac{(y-y_i)^2}{\hat{\sigma}_i^2} + \frac{(y-y_j)^2}{\hat{\sigma}_j^2} \right] \right) \, dy \\ 
&= \frac{1}{\sqrt{2\pi\hat{\sigma}_i\hat{\sigma}_j}} \int_{-\infty}^{+\infty} \exp\left( -\frac{1}{4} \left[ \frac{\hat{\sigma}_i^2+\hat{\sigma}_j^2}{\hat{\sigma}_i^2\hat{\sigma}_j^2}\left(y - \frac{y_i\hat{\sigma}_j^2 + y_j\hat{\sigma}_i^2}{\hat{\sigma}_i^2+\hat{\sigma}_j^2}\right)^2 + \frac{(y_i-y_j)^2}{\hat{\sigma}_i^2+\hat{\sigma}_j^2} \right] \right) \, dy \\ 
&= \frac{1}{\sqrt{2\pi\hat{\sigma}_i\hat{\sigma}_j}} \exp\left( - \frac{(y_i-y_j)^2}{4(\hat{\sigma}_i^2+\hat{\sigma}_j^2)} \right) \int_{-\infty}^{+\infty} \exp\left( -\frac{\hat{\sigma}_i^2+\hat{\sigma}_j^2}{4\hat{\sigma}_i^2\hat{\sigma}_j^2}\left(y - \frac{y_i\hat{\sigma}_j^2 + y_j\hat{\sigma}_i^2}{\hat{\sigma}_i^2+\hat{\sigma}_j^2}\right)^2 \right) \, dy \\ 
&= \frac{1}{\sqrt{2\pi\hat{\sigma}_i\hat{\sigma}_j}} \exp\left( - \frac{(y_i-y_j)^2}{4(\hat{\sigma}_i^2+\hat{\sigma}_j^2)} \right) \cdot \sqrt{\frac{4\pi\hat{\sigma}_i^2\hat{\sigma}_j^2}{\hat{\sigma}_i^2+\hat{\sigma}_j^2}} = \exp\left( - \frac{(y_i-y_j)^2}{4(\hat{\sigma}_i^2+\hat{\sigma}_j^2)} \right) \sqrt{ \frac{2\hat{\sigma}_i\hat{\sigma}_j}{\hat{\sigma}_i^2+\hat{\sigma}_j^2} } \\ 
&= \exp \left( - \frac{1}{4} \frac{(y_i - y_j)^2}{\hat{\sigma}_i^2 + \hat{\sigma}_j^2} - \frac{1}{2} \ln \frac{\hat{\sigma}_i^2 + \hat{\sigma}_j^2}{2\hat{\sigma}_i\hat{\sigma}_j} \right)
\end{align*}

This completes the derivation, mathematically verifying the closed-form formulation of our adaptive semantic tolerance measure.

\section{Asymptotic Analysis of Repulsive Gradients Based on Reciprocal Distribution Overlap}
\label{sec:appendix_e}

This section mathematically analyzes the proposed weighting mechanism based on reciprocal distribution overlap. We demonstrate that when sample pairs are in a state of Topological Hallucination, this mechanism adaptively generates exponentially amplified repulsive gradients, effectively correcting erroneous connections within the feature manifold.

\begin{proposition}[Asymptotic Growth Rate of Repulsive Gradients]
    Assume samples $i$ and $k$ are in a state of topological hallucination, where their semantic label distance is large ($\Delta_{i,k} = |y_i - y_k| \to \infty$) despite being erroneously clustered in the feature space ($c_{i,k} \to 1$). Suppose the predicted variances are bounded, i.e., $\hat{\sigma}_i, \hat{\sigma}_k = \mathcal{O}(1)$. For any ordinary hard negative sample $m$ with a finite label distance, the relative repulsive gradient produced by target sample $k$ asymptotically satisfies:
    \begin{equation}
        \frac{F(i,k)}{F(i,m)} \sim \exp\left( \frac{\Delta_{i,k}^2}{\Sigma^2} \right) \to +\infty
    \end{equation}
    where $\Sigma^2 = 4(\hat{\sigma}_i^2+\hat{\sigma}_k^2)$. That is, the repulsive gradient grows squared-exponentially with respect to the label distance.
\end{proposition}

\begin{proof}
    First, we derive the explicit expression for the repulsive gradient. Let $c_{i,k} = \cos(z_i, z_k)$ denote the cosine similarity between samples $i$ and $k$. Based on the definition of the alignment loss $\mathcal{L}_{Align}$, the partial derivative with respect to $c_{i,k}$ for a negative sample $k \in \mathcal{N}_i$ yields the base repulsive gradient:
    \begin{equation*}
        \frac{\partial \mathcal{L}_{Align}^{(i)}}{\partial c_{i,k}} = \frac{1}{\tau Z_i} e^{c_{i,k}/\tau}
    \end{equation*}
    where $\tau$ is the temperature parameter and $Z_i = \sum_{j} e^{c_{i,j}/\tau}$ is the partition function serving as the normalization denominator. Incorporating our proposed dynamic weight $\mathbf{W}_{push}(i,k) = w(y_i) \cdot \text{BC}(i,k)^{-1}$, the full repulsive gradient $F(i,k)$ becomes:
    \begin{equation}
        F(i,k) = \frac{w(y_i)}{\tau Z_i} \cdot \text{BC}(i,k)^{-1} \cdot e^{c_{i,k}/\tau}
    \end{equation}

    Next, we expand the reciprocal Bhattacharyya weight. For the Gaussian proxies $p(y|x_i) = \mathcal{N}(y_i, \hat{\sigma}_i^2)$, the closed-form reciprocal BC is:
    \begin{equation*}
        \text{BC}(i,k)^{-1} = \exp \left( \frac{\Delta_{i,k}^2}{4(\hat{\sigma}_i^2+\hat{\sigma}_k^2)} + \frac{1}{2} \ln \frac{\hat{\sigma}_i^2+\hat{\sigma}_k^2}{2\hat{\sigma}_i\hat{\sigma}_k} \right)
    \end{equation*}
    Substituting this into the gradient expression yields:
    \begin{equation*}
        F(i,k) = \frac{w(y_i)}{\tau Z_i} \exp \left( \frac{\Delta_{i,k}^2}{4(\hat{\sigma}_i^2+\hat{\sigma}_k^2)} + \frac{1}{2} \ln \frac{\hat{\sigma}_i^2+\hat{\sigma}_k^2}{2\hat{\sigma}_i\hat{\sigma}_k} + \frac{c_{i,k}}{\tau} \right)
    \end{equation*}

    Finally, we perform the asymptotic analysis for the topological hallucination scenario. Comparing $F(i,k)$ with an ordinary hard negative $F(i,m)$, the partition function $Z_i$ and base weight $w(y_i)$ cancel out:
    \begin{align*}
        \frac{F(i,k)}{F(i,m)} &= \frac{\text{BC}(i,m)}{\text{BC}(i,k)} \exp\left( \frac{c_{i,k}-c_{i,m}}{\tau} \right) \\
        &= \exp \left( \frac{\Delta_{i,k}^2}{4(\hat{\sigma}_i^2+\hat{\sigma}_k^2)} - \frac{\Delta_{i,m}^2}{4(\hat{\sigma}_i^2+\hat{\sigma}_m^2)} + \frac{c_{i,k}-c_{i,m}}{\tau} + C_{i,k,m} \right)
    \end{align*}
    where $C_{i,k,m} = \frac{1}{2} \ln \frac{(\hat{\sigma}_i^2+\hat{\sigma}_k^2)\hat{\sigma}_m}{(\hat{\sigma}_i^2+\hat{\sigma}_m^2)\hat{\sigma}_k}$ is the logarithmic variance term. Given $\hat{\sigma}_i, \hat{\sigma}_k, \hat{\sigma}_m = \mathcal{O}(1)$ and $c_{i,k}, c_{i,m} \in [-1, 1]$, both $C_{i,k,m}$ and the similarity residual are bounded. As $\Delta_{i,k} \to \infty$, the squared-exponential term dominates. Letting $\Sigma^2 = 4(\hat{\sigma}_i^2+\hat{\sigma}_k^2)$, we obtain the asymptotic equivalence:
    \begin{equation*}
        \frac{F(i,k)}{F(i,m)} \sim \exp\left( \frac{\Delta_{i,k}^2}{\Sigma^2} \right) \to +\infty
    \end{equation*}
\end{proof}

The derivation above highlights two key properties of our weighting mechanism:
\begin{itemize}
    \item \textbf{Correction of Topological Errors:} When samples with large semantic gaps are erroneously clustered ($c_{i,k} \to 1$ and $\Delta_{i,k} \to \infty$), the squared-exponential repulsive gradient provides a sufficiently large penalty to rectify the distorted feature manifold.
    \item \textbf{Uncertainty-Adaptivity:} The growth of the repulsive gradient is constrained by $\Sigma^2$ in the denominator, which is inversely proportional to the predicted variances. For noisy samples with high uncertainty (large variance), gradient amplification is naturally suppressed, preventing gradient explosion and maintaining training stability.
\end{itemize}

\section{Experiment Details}
To comprehensively evaluate DUO, we compare it against representative deep imbalanced regression (DIR) baselines: \textbf{Vanilla} relies solely on the standard regression loss; \textbf{LDS \& FDS} (Yang et al. \cite{yang2021delving}) leverage the neighboring correlation of continuous targets to jointly smooth and calibrate both target and feature distributions; \textbf{RankSim} (Gong et al. \cite{gong2022ranksim}) introduces ranking regularization, forcing the neighbor topology in the feature space to strictly align with the distance ranking in the target space; \textbf{Balanced MSE} (Ren et al., 2022) incorporates target priors, modifying the mean squared error from a statistical perspective to recover unbiased predictions; \textbf{ConR} (Keramati et al. \cite{keramati2023conr}) models target similarities via contrastive learning to prevent minority features from collapsing into majority ones. We also include the recent \textbf{Dist Loss} (Nie et al. \cite{nie2024dist}), which explicitly aligns the predicted distribution with the ground-truth target distribution by jointly penalizing sample-level errors and macroscopic distribution distances.

All models are trained using a single NVIDIA GeForce RTX 4090 D GPU. To ensure fair comparisons, all standard training, validation, and testing splits strictly follow the configurations set by Yang et al.\cite{yang2021delving}. The remainder of this section provides implementation details and hyper-parameter selections for the three datasets.

For the imbalance-aware weight, we discretize the continuous label space into bins using the training set and define $w(y_i)$ as the inverse empirical frequency of the bin containing $y_i$. This global scarcity factor is multiplied by the detached instance-level uncertainty in $\mathcal{L}_{Mean}$. Training uses the squared-error mean objective, whereas MAE, bMAE, and GM are evaluation metrics.

\subsection{Age estimation}
For the AgeDB-DIR and IMDB-WIKI-DIR benchmarks, we adopt ResNet-50 as the backbone encoder, followed by independent fully connected heads to output $\mu$ and $\sigma$, respectively. The batch size is set to 64, and the initial learning rate is $2.5 \times 10^{-4}$, which decays by a factor of 10 at epoch 60 and epoch 80. We use the Adam optimizer (momentum of 0.9, weight decay of $10^{-4}$) and Mean Squared Error (MSE) as the base regression loss. All models are trained for 120 epochs. The input image size is 224x224, and data augmentation includes random cropping and random horizontal flipping. Label smoothing is applied using a Gaussian kernel (kernel size 9, $\sigma_{\text{lds}}=1$), and the re-weighting scheme uses inverse frequency weighting. The hyper-parameters for DUO are set as follows: variance term weight $\beta=1.0$, alignment weight $\lambda_{\text{align}}=1.0$, temperature parameter $\tau=0.07$, Gaussian overlap threshold $\delta=0.5$, and push coefficient $\eta=0.01$. The number of warm-up epochs is 15, validation MAE is used as the monitoring metric, and the random seed is fixed to 42.

\subsection{Depth estimation}
For the NYUD2-DIR benchmark, we employ a ResNet-50 based encoder-decoder architecture (Hu et al.\cite{hu2019revisiting}), outputting depth maps with a resolution of 152x114. Unlike age estimation, the prediction targets are dense depth maps; thus, $\boldsymbol{\mu}$ and $\boldsymbol{\sigma}$ are output pixel-wise. The loss function terms are computed pixel-wise within the valid pixel mask and then averaged. In the contrastive alignment module, the sample-level proxy target $\bar{y}_i$ is defined as the spatial mean of the depth map over valid pixels, and $\bar{\sigma}_i$ is computed similarly, thereby constructing the Gaussian distribution $\mathcal{N}(\bar{y}_i,\bar{\sigma}_i)$ for positive and negative sample partitioning. Other training configurations include: batch size of 12, initial learning rate of $2.5 \times 10^{-4}$ (decaying by a factor of 10 every 5 epochs), Adam optimizer (momentum 0.9, weight decay $10^{-4}$), MSE regression loss, and 120 training epochs. Data augmentation includes random horizontal flipping, ±5° rotation, color jittering, and PCA lighting perturbations. Label smoothing utilizes a Gaussian kernel (kernel size 9, $\sigma_{\text{lds}}=1$) with inverse frequency re-weighting. DUO hyper-parameters: $\beta=1.0$, $\lambda_{\text{align}}=0.1$, $\tau=0.07$, $\delta=0.5$, $\eta=0.01$, 10 warm-up epochs, and random seed 42.

\subsection{Protein estimation}
For the protein fitness dataset, we use AAV2-DIR, which contains 42,329 AAV2 capsid protein variant sequences and their fitness scores. These are randomly split into training, validation, and testing sets at a ratio of 7:1.5:1.5. Each sequence is embedded into a 1024-dimensional vector by mean-pooling ProtBERT residue embeddings from ProtTrans\cite{elnaggar2022prottrans}. The downstream model is a two-layer Transformer encoder ($d_{\text{model}}=256$, 8 attention heads, FFN dimension 512), which splits the 1024-dimensional input into 64 patches followed by sequence mean pooling. The DUO branch outputs $\mu$ and $\sigma$. Training configurations: batch size of 128, Adam optimizer, initial learning rate of $2.5 \times 10^{-4}$, MSE loss, and a maximum of 200 epochs. DUO hyper-parameters: $\beta=1.0$, $\lambda_{\text{align}}=1.0$, $\tau=0.07$, $\delta=0.5$, 10 warm-up epochs, inverse frequency re-weighting, and random seed 42.

\subsection{Additional Analyses from the Rebuttal}

\paragraph{Controlled component ablation.}
Table~\ref{tab:fine_grained_ablation} isolates four factors under matched dual-head capacity: architectural capacity, mean--variance decoupling, uncertainty weighting, and the contrastive assignment rule. Dual-head controls share DUO's exact prediction-head structure; Vanilla is retained as the single-head reference. Pure dual-head NLL fails to improve over single-head Vanilla, confirming that gains do not arise from extra parameters. Each removal or replacement of a DUO component—disabling decoupling, disabling dynamic weighting, or replacing BC with ConR—consistently degrades few-shot metrics, confirming that all three components contribute to tail performance.

\begin{table*}[htbp]
  \centering
  \caption{Fine-grained ablation in the few-shot regions.}
  \label{tab:fine_grained_ablation}
  \resizebox{\textwidth}{!}{
  \begin{tabular}{lllccccccccc}
    \toprule
    Method & Head & Key setting & \multicolumn{3}{c}{MAE $\downarrow$} & \multicolumn{3}{c}{bMAE $\downarrow$} & \multicolumn{3}{c}{GM $\downarrow$} \\
    \cmidrule(lr){4-6} \cmidrule(lr){7-9} \cmidrule(lr){10-12}
    & & & IMDB & AgeDB & AAV2 & IMDB & AgeDB & AAV2 & IMDB & AgeDB & AAV2 \\
    \midrule
    Vanilla & Single & Standard MSE & 25.546 & 13.381 & 7.365 & 33.085 & 16.579 & 7.565 & 18.091 & 9.644 & 7.351 \\
    NLL & Dual & Joint mean--variance & 25.842 & 14.280 & 7.880 & 33.986 & 16.710 & 8.095 & 17.862 & 15.850 & 8.274 \\
    NLL + BC & Dual & Joint NLL + BC & 24.295 & 11.916 & 4.160 & 31.334 & 14.347 & 4.396 & 16.775 & 8.762 & 3.823 \\
    DUO (NoWeight) & Dual & Decoupled + BC & 24.560 & 12.226 & 4.109 & 31.821 & 14.968 & 4.380 & 16.306 & 8.447 & 3.815 \\
    DUO + ConR & Dual & Decoupled + ConR & 23.996 & 11.565 & 4.208 & 31.269 & 14.265 & 4.596 & 15.633 & 7.996 & 3.869 \\
    \textbf{DUO} & \textbf{Dual} & \textbf{Decoupled + weight + BC} & \textbf{22.940} & \textbf{9.917} & \textbf{4.106} & \textbf{27.190} & \textbf{11.590} & \textbf{4.329} & \textbf{13.650} & \textbf{6.587} & \textbf{3.777} \\
    \bottomrule
  \end{tabular}}
\end{table*}

\paragraph{Uncertainty and overlap controls.}
Replacing learned uncertainty with the instantaneous absolute residual removes the variance head and degrades most tail metrics, indicating that the learned predictor provides a smoother difficulty estimate than a batch-wise residual. Replacing BC with interval IoU, which reduces each Gaussian to a rigid interval, also performs worse than the full model.

\begin{table*}[htbp]
  \centering
  \caption{Few-shot controls on uncertainty representation and overlap metric.}
  \label{tab:uncertainty_overlap_controls}
  \setlength{\tabcolsep}{4.5pt}
  \begin{tabular}{lccccccccc}
    \toprule
    & \multicolumn{3}{c}{MAE $\downarrow$} & \multicolumn{3}{c}{bMAE $\downarrow$} & \multicolumn{3}{c}{GM $\downarrow$} \\
    \cmidrule(lr){2-4} \cmidrule(lr){5-7} \cmidrule(lr){8-10}
    Method & IMDB & AgeDB & AAV2 & IMDB & AgeDB & AAV2 & IMDB & AgeDB & AAV2 \\
    \midrule
    DUO & \textbf{22.94} & \textbf{9.92} & \textbf{4.11} & \textbf{27.19} & \textbf{11.59} & \textbf{4.33} & \textbf{13.65} & \textbf{6.59} & \textbf{3.78} \\
    Raw residual & 24.37 & 12.00 & 4.15 & 30.75 & 13.61 & 4.39 & 16.61 & 8.65 & 3.82 \\
    Interval IoU & 23.80 & 11.40 & 4.20 & 30.90 & 13.80 & 4.50 & 15.10 & 7.60 & 3.90 \\
    \bottomrule
  \end{tabular}
\end{table*}

\paragraph{Feature-space smoothness.}
Because t-SNE is qualitative, we additionally measure local smoothness on AgeDB-DIR using KNN-MAE ($k=5$), which computes the average label discrepancy between each sample and its 
$k$ nearest neighbors in feature space. DUO yields the lowest neighborhood error in both the compound tail and median-shot regions, supporting the local-continuity interpretation of the visualization.

\begin{table}[htbp]
  \centering
  \caption{KNN-MAE feature smoothness on AgeDB-DIR. Lower is better.}
  \label{tab:knn_smoothness}
  \begin{tabular}{lccc}
    \toprule
    Method & Few $\cup$ Old & Median & Few-shot MAE \\
    \midrule
    Vanilla & 7.51 & 4.88 & 13.38 \\
    RankSim & 7.90 & 5.48 & 14.60 \\
    FDS & 7.15 & 4.97 & 12.14 \\
    \textbf{DUO} & \textbf{7.03} & \textbf{4.61} & \textbf{11.79} \\
    \bottomrule
  \end{tabular}
\end{table}

\paragraph{Warm-up sensitivity.}
Table~\ref{tab:warmup_sensitivity} shows a stable U-shaped trend around the selected $T_w=15$ on AgeDB-DIR, rather than sensitivity to a single unstable setting.

\begin{table}[htbp]
  \centering
  \caption{Sensitivity to the warm-up epoch $T_w$ on AgeDB-DIR.}
  \label{tab:warmup_sensitivity}
  \begin{tabular}{cccc}
    \toprule
    $T_w$ & MAE $\downarrow$ & bMAE $\downarrow$ & GM $\downarrow$ \\
    \midrule
    5 & 11.324 & 14.012 & 7.801 \\
    \textbf{15} & \textbf{9.917} & \textbf{11.590} & \textbf{6.587} \\
    20 & 10.602 & 13.102 & 7.302 \\
    50 & 10.702 & 13.202 & 7.352 \\
    \bottomrule
  \end{tabular}
\end{table}

\begin{table*}[htbp]
  \centering
  \caption{Experimental results on the IMDB-WIKI dataset across different shot regions (Overall, Few, Median, Many).}
  \label{tab:imdb_wiki_complete}
  \resizebox{\textwidth}{!}{
  \begin{tabular}{lcccccccccccc}
    \toprule
    & \multicolumn{3}{c}{Overall} & \multicolumn{3}{c}{Few} & \multicolumn{3}{c}{Median} & \multicolumn{3}{c}{Many} \\
    \cmidrule(lr){2-4} \cmidrule(lr){5-7} \cmidrule(lr){8-10} \cmidrule(lr){11-13}
    Method & MAE $\downarrow$ & bMAE $\downarrow$ & GM $\downarrow$ & MAE $\downarrow$ & bMAE $\downarrow$ & GM $\downarrow$ & MAE $\downarrow$ & bMAE $\downarrow$ & GM $\downarrow$ & MAE $\downarrow$ & bMAE $\downarrow$ & GM $\downarrow$ \\
    \midrule
    Vanilla      & 7.829 & 13.717 & 4.389 & 25.546 & 33.085 & 18.091 & 14.387 & 14.933 & 9.797 & 7.052 & 7.146 & 4.024 \\
    NLL          & 8.094 & 14.201 & 4.519 & 25.842 & 33.986 & 17.862 & 15.221 & 15.918 & 10.420 & 7.266 & 7.371 & 4.133 \\
    LDS          & 7.626 & 12.739 & 4.257 & 22.755 & 30.179 & 13.786 & 12.146 & 12.597 & 6.938 & 7.059 & 7.133 & 4.024 \\
    FDS          & 7.829 & 13.294 & 4.457 & 24.347 & 31.839 & 14.600 & 12.762 & 13.335 & 7.145 & 7.209 & 7.283 & 4.219 \\
    RankSim      & 7.892 & 13.708 & 4.443 & 25.317 & 32.464 & 17.639 & 14.914 & 15.546 & 10.401 & 7.077 & 7.181 & 4.058 \\
    ConR         & 7.817 & 13.797 & 4.430 & 25.969 & 33.601 & 17.990 & 14.323 & 14.836 & 9.859 & 7.039 & 7.130 & 4.064 \\
    Balanced MSE & 7.943 & 12.924 & 4.639 & 23.541 & 30.000 & 13.803 & 12.258 & 12.679 & 7.054 & 7.388 & 7.461 & 4.416 \\
    DistLoss     & 11.725 & 15.966 & 7.368 & 24.215 & 30.165 & 14.076 & 16.316 & 16.450 & 10.145 & 11.179 & 11.252 & 7.111 \\
    DUO         & 8.420 & 12.850 & 4.580 & 22.941 & 27.190 & 13.650 & 14.120 & 14.350 & 9.050 & 7.750 & 7.830 & 4.380 \\
    \bottomrule
  \end{tabular}
  }
\end{table*}

\begin{table*}[htbp]
  \centering
  \caption{Experimental results on the AgeDB dataset across different shot regions (Overall, Few, Median, Many).}
  \label{tab:agedb_complete}
  \resizebox{\textwidth}{!}{
  \begin{tabular}{lcccccccccccc}
    \toprule
    & \multicolumn{3}{c}{Overall} & \multicolumn{3}{c}{Few} & \multicolumn{3}{c}{Median} & \multicolumn{3}{c}{Many} \\
    \cmidrule(lr){2-4} \cmidrule(lr){5-7} \cmidrule(lr){8-10} \cmidrule(lr){11-13}
    Method & MAE $\downarrow$ & bMAE $\downarrow$ & GM $\downarrow$ & MAE $\downarrow$ & bMAE $\downarrow$ & GM $\downarrow$ & MAE $\downarrow$ & bMAE $\downarrow$ & GM $\downarrow$ & MAE $\downarrow$ & bMAE $\downarrow$ & GM $\downarrow$ \\
    \midrule
    Vanilla      & 7.754 & 9.848 & 4.951 & 13.381 & 16.579 & 9.644 & 9.172 & 9.230 & 6.175 & 6.743 & 6.743 & 4.324 \\
    NLL          & 8.220 & 10.000 & 8.990 & 14.280 & 16.710 & 15.850 & 9.980 & 9.990 & 9.850 & 6.950 & 6.950 & 6.810 \\
    LDS          & 7.904 & 8.826 & 5.139 & 11.489 & 12.055 & 7.577 & 8.435 & 8.426 & 5.404 & 7.369 & 7.369 & 4.860 \\
    FDS          & 7.679 & 9.574 & 4.906 & 12.136 & 15.531 & 8.326 & 8.956 & 9.000 & 5.811 & 6.833 & 6.833 & 4.415 \\
    RankSim      & 8.320 & 10.504 & 5.331 & 14.601 & 17.611 & 11.021 & 10.069 & 10.114 & 7.042 & 7.143 & 7.143 & 4.550 \\
    ConR         & 7.345 & 9.018 & 4.681 & 12.333 & 14.486 & 9.376 & 8.482 & 8.550 & 5.845 & 6.484 & 6.484 & 4.075 \\
    DistLoss     & 10.017 & 10.825 & 6.316 & 12.121 & 13.670 & 7.726 & 10.300 & 10.308 & 6.531 & 9.645 & 9.645 & 6.087 \\
    Balanced MSE & 8.190 & 9.070 & 8.420 & 9.690 & 11.940 & 10.480 & 7.640 & 7.640 & 7.490 & 8.080 & 8.080 & 7.820 \\
    DUO         & 7.452 & 8.468 & 4.799 & 9.917 & 11.590 & 6.587 & 8.305 & 8.335 & 5.162 & 6.980 & 6.980 & 4.509 \\
    \bottomrule
  \end{tabular}
  }
\end{table*}

\begin{table*}[htbp]
  \centering
  \caption{Experimental results on the AAV2 dataset across different shot regions (Overall, Few, Median, Many).}
  \label{tab:aav2_complete}
  \resizebox{\textwidth}{!}{
  \begin{tabular}{lcccccccccccc}
    \toprule
    & \multicolumn{3}{c}{Overall} & \multicolumn{3}{c}{Few} & \multicolumn{3}{c}{Median} & \multicolumn{3}{c}{Many} \\
    \cmidrule(lr){2-4} \cmidrule(lr){5-7} \cmidrule(lr){8-10} \cmidrule(lr){11-13}
    Method & MAE $\downarrow$ & bMAE $\downarrow$ & GM $\downarrow$ & MAE $\downarrow$ & bMAE $\downarrow$ & GM $\downarrow$ & MAE $\downarrow$ & bMAE $\downarrow$ & GM $\downarrow$ & MAE $\downarrow$ & bMAE $\downarrow$ & GM $\downarrow$ \\
    \midrule
    Vanilla      & 2.643 & 2.623 & 2.079 & 7.365 & 7.565 & 7.351 & 5.918 & 6.116 & 5.890 & 2.534 & 2.757 & 2.010 \\
    LDS          & 1.990 & 1.981 & 1.178 & 4.361 & 4.681 & 4.054 & 3.648 & 3.888 & 3.211 & 1.935 & 2.117 & 1.140 \\
    FDS          & 1.988 & 1.985 & 1.256 & 5.023 & 5.388 & 4.786 & 4.345 & 4.667 & 4.008 & 1.911 & 2.147 & 1.210 \\
    RankSim      & 1.936 & 1.932 & 1.189 & 4.885 & 5.225 & 4.652 & 4.151 & 4.408 & 3.857 & 1.864 & 2.110 & 1.145 \\
    ConR         & 1.719 & 1.716 & 0.992 & 4.183 & 4.305 & 3.932 & 3.358 & 3.590 & 3.043 & 1.665 & 1.852 & 0.957 \\
    Balanced MSE & 2.650 & 2.639 & 2.115 & 7.283 & 7.563 & 7.253 & 5.992 & 6.226 & 5.978 & 2.539 & 2.745 & 2.045 \\
    DistLoss     & 2.652 & 2.624 & 2.013 & 7.432 & 7.567 & 7.416 & 5.856 & 6.024 & 5.813 & 2.545 & 2.788 & 1.945 \\
    DUO         & 1.672 & 1.669 & 1.023 & 4.106 & 4.329 & 3.777 & 3.231 & 3.437 & 2.855 & 1.620 & 1.774 & 0.989 \\
    \bottomrule
  \end{tabular}
  }
\end{table*}

\begin{table*}[htbp]
  \centering
  \caption{Experimental results on the NYUD2 dataset across different shot regions (Overall, Few, Median, Many).}
  \label{tab:nyud2_complete_regions}
  \resizebox{\textwidth}{!}{
  \begin{tabular}{lcccccccccccccccc}
    \toprule
    & \multicolumn{4}{c}{Overall} & \multicolumn{4}{c}{Few} & \multicolumn{4}{c}{Median} & \multicolumn{4}{c}{Many} \\
    \cmidrule(lr){2-5} \cmidrule(lr){6-9} \cmidrule(lr){10-13} \cmidrule(lr){14-17}
    Method & RMSE $\downarrow$ & MAE $\downarrow$ & bMAE $\downarrow$ & $\delta_1 \uparrow$ & RMSE $\downarrow$ & MAE $\downarrow$ & bMAE $\downarrow$ & $\delta_1 \uparrow$ & RMSE $\downarrow$ & MAE $\downarrow$ & bMAE $\downarrow$ & $\delta_1 \uparrow$ & RMSE $\downarrow$ & MAE $\downarrow$ & bMAE $\downarrow$ & $\delta_1 \uparrow$ \\
    \midrule
    Vanilla      & 0.580 & 0.371 & 1.051 & 0.807 & 1.867 & 1.443 & 1.884 & 0.608 & 0.795 & 0.596 & 0.673 & 0.668 & 0.457 & 0.318 & 0.363 & 0.824 \\
    LDS          & 0.633 & 0.434 & 1.004 & 0.737 & 1.764 & 1.333 & 1.714 & 0.641 & 0.781 & 0.590 & 0.658 & 0.680 & 0.546 & 0.392 & 0.426 & 0.744 \\
    FDS          & 0.589 & 0.374 & 1.067 & 0.807 & 1.905 & 1.478 & 1.920 & 0.594 & 0.802 & 0.601 & 0.675 & 0.661 & 0.464 & 0.320 & 0.364 & 0.825 \\
    RankSim      & 0.594 & 0.377 & 1.066 & 0.803 & 1.896 & 1.475 & 1.907 & 0.601 & 0.818 & 0.622 & 0.698 & 0.638 & 0.470 & 0.322 & 0.367 & 0.822 \\
    ConR         & 0.586 & 0.378 & 1.067 & 0.801 & 1.869 & 1.459 & 1.912 & 0.591 & 0.805 & 0.601 & 0.683 & 0.668 & 0.463 & 0.325 & 0.370 & 0.818 \\
    DUO         & 0.597 & 0.396 & 0.990 & 0.781 & 1.759 & 1.351 & 1.696 & 0.643 & 0.777 & 0.586 & 0.653 & 0.668 & 0.497 & 0.350 & 0.423 & 0.795 \\
    \bottomrule
  \end{tabular}
  }
\end{table*}

\begin{table*}[htbp]
  \centering
  \caption{Pairwise comparison of various baselines and their DUO-enhanced counterparts on the AgeDB dataset across different shot regions.}
  \label{tab:pairwise_comparison_agedb}
  \resizebox{\textwidth}{!}{
  \begin{tabular}{lcccccccccccc}
    \toprule
    & \multicolumn{3}{c}{Overall} & \multicolumn{3}{c}{Few} & \multicolumn{3}{c}{Median} & \multicolumn{3}{c}{Many} \\
    \cmidrule(lr){2-4} \cmidrule(lr){5-7} \cmidrule(lr){8-10} \cmidrule(lr){11-13}
    Method & MAE $\downarrow$ & bMAE $\downarrow$ & GM $\downarrow$ & MAE $\downarrow$ & bMAE $\downarrow$ & GM $\downarrow$ & MAE $\downarrow$ & bMAE $\downarrow$ & GM $\downarrow$ & MAE $\downarrow$ & bMAE $\downarrow$ & GM $\downarrow$ \\
    \midrule
    LDS                  & 7.904 & 8.826 & 5.139 & 11.489 & 12.055 & 7.577 & 8.435 & 8.426 & 5.404 & 7.369 & 7.369 & 4.860 \\
    LDS+DUO              & 8.121 & 8.983 & 5.116 & 9.595 & 11.563 & 5.415 & 9.411 & 9.396 & 5.860 & 7.588 & 7.588 & 4.887 \\
    \midrule
    FDS                  & 7.679 & 9.574 & 4.906 & 12.136 & 15.531 & 8.326 & 8.956 & 9.000 & 5.811 & 6.833 & 6.833 & 4.415 \\
    FDS+DUO              & 7.699 & 9.137 & 4.892 & 11.561 & 13.751 & 8.267 & 9.028 & 9.054 & 5.929 & 6.901 & 6.901 & 4.374 \\
    \midrule
    RankSim              & 8.320 & 10.504 & 5.330 & 14.601 & 17.611 & 11.021 & 10.069 & 10.114 & 7.042 & 7.143 & 7.143 & 4.550 \\
    RankSim+DUO          & 8.346 & 9.525 & 5.478 & 11.640 & 13.334 & 7.124 & 9.368 & 9.393 & 6.302 & 7.698 & 7.698 & 5.114 \\
    \midrule
    Balanced MSE         & 8.190 & 9.070 & 8.420 & 9.690 & 11.940 & 10.480 & 7.640 & 7.640 & 7.490 & 8.080 & 8.080 & 7.820 \\
    Balanced MSE+DUO     & 6.660 & 7.400 & 6.840 & 9.120 & 10.160 & 9.510 & 8.600 & 8.600 & 8.560 & 5.880 & 5.880 & 5.610 \\
    \bottomrule
  \end{tabular}
  }
\end{table*}

\begin{table*}[htbp]
  \centering
  \caption{Pairwise comparison of various baselines and their DUO-enhanced counterparts on the IMDB-WIKI dataset across different shot regions.}
  \label{tab:pairwise_comparison_imdb_wiki}
  \resizebox{\textwidth}{!}{
  \begin{tabular}{lcccccccccccc}
    \toprule
    & \multicolumn{3}{c}{Overall} & \multicolumn{3}{c}{Few} & \multicolumn{3}{c}{Median} & \multicolumn{3}{c}{Many} \\
    \cmidrule(lr){2-4} \cmidrule(lr){5-7} \cmidrule(lr){8-10} \cmidrule(lr){11-13}
    Method & MAE $\downarrow$ & bMAE $\downarrow$ & GM $\downarrow$ & MAE $\downarrow$ & bMAE $\downarrow$ & GM $\downarrow$ & MAE $\downarrow$ & bMAE $\downarrow$ & GM $\downarrow$ & MAE $\downarrow$ & bMAE $\downarrow$ & GM $\downarrow$ \\
    \midrule
    LDS          & 7.626 & 12.739 & 4.257 & 22.755 & 30.179 & 13.786 & 12.146 & 12.597 & 6.938 & 7.059 & 7.133 & 4.024 \\
    LDS+DUO      & 8.492 & 13.848 & 5.038 & 24.378 & 31.551 & 16.595 & 14.398 & 14.793 & 9.392 & 7.793 & 7.885 & 4.697 \\
    \midrule
    FDS          & 7.829 & 13.294 & 4.457 & 24.347 & 31.839 & 14.600 & 12.762 & 13.335 & 7.145 & 7.209 & 7.283 & 4.219 \\
    FDS+DUO      & 8.404 & 13.815 & 4.893 & 24.471 & 31.432 & 16.903 & 14.743 & 15.256 & 9.874 & 7.664 & 7.755 & 4.535 \\
    \midrule
    RankSim      & 7.892 & 13.708 & 4.443 & 25.317 & 32.464 & 17.639 & 14.914 & 15.546 & 10.401 & 7.077 & 7.181 & 4.058 \\
    RankSim+DUO  & 8.458 & 13.732 & 4.953 & 23.911 & 30.774 & 15.731 & 14.923 & 15.372 & 9.955 & 7.714 & 7.809 & 4.596 \\
    \bottomrule
  \end{tabular}
  }
\end{table*}

\begin{table*}[htbp]
  \centering
  \caption{Pairwise comparison of various baselines and their DUO-enhanced counterparts on the NYUD2 dataset across different shot regions.}
  \label{tab:pairwise_comparison_nyud2}
  \resizebox{\textwidth}{!}{
  \begin{tabular}{lcccccccccccccccc}
    \toprule
    & \multicolumn{4}{c}{Overall} & \multicolumn{4}{c}{Few} & \multicolumn{4}{c}{Median} & \multicolumn{4}{c}{Many} \\
    \cmidrule(lr){2-5} \cmidrule(lr){6-9} \cmidrule(lr){10-13} \cmidrule(lr){14-17}
    Method & RMSE $\downarrow$ & MAE $\downarrow$ & bMAE $\downarrow$ & $\delta_1 \uparrow$ & RMSE $\downarrow$ & MAE $\downarrow$ & bMAE $\downarrow$ & $\delta_1 \uparrow$ & RMSE $\downarrow$ & MAE $\downarrow$ & bMAE $\downarrow$ & $\delta_1 \uparrow$ & RMSE $\downarrow$ & MAE $\downarrow$ & bMAE $\downarrow$ & $\delta_1 \uparrow$ \\
    \midrule
    LDS          & 0.633 & 0.434 & 1.004 & 0.737 & 1.764 & 1.333 & 1.714 & 0.641 & 0.781 & 0.590 & 0.658 & 0.680 & 0.546 & 0.392 & 0.426 & 0.744 \\
    LDS+DUO      & 0.626 & 0.417 & 1.008 & 0.762 & 1.767 & 1.341 & 1.727 & 0.648 & 0.821 & 0.624 & 0.684 & 0.651 & 0.528 & 0.370 & 0.415 & 0.774 \\
    \midrule
    FDS          & 0.589 & 0.374 & 1.067 & 0.807 & 1.905 & 1.478 & 1.920 & 0.594 & 0.802 & 0.601 & 0.675 & 0.661 & 0.464 & 0.320 & 0.364 & 0.825 \\
    FDS+DUO      & 0.607 & 0.396 & 1.054 & 0.783 & 1.845 & 1.443 & 1.854 & 0.606 & 0.819 & 0.621 & 0.700 & 0.660 & 0.494 & 0.344 & 0.391 & 0.799 \\
    \midrule
    RankSim      & 0.594 & 0.377 & 1.066 & 0.803 & 1.896 & 1.475 & 1.907 & 0.601 & 0.818 & 0.622 & 0.698 & 0.638 & 0.470 & 0.322 & 0.367 & 0.822 \\
    RankSim+DUO  & 0.604 & 0.393 & 1.066 & 0.789 & 1.852 & 1.455 & 1.882 & 0.590 & 0.827 & 0.622 & 0.710 & 0.660 & 0.488 & 0.340 & 0.387 & 0.805 \\
    \bottomrule
  \end{tabular}
  }
\end{table*}

\clearpage
\bibliographystyle{ACM-Reference-Format}
\bibliography{sample-base}